\documentclass[letterpaper]{article} 
\usepackage{aaai25}  
\usepackage{times}  
\usepackage{helvet}  
\usepackage{courier}  
\usepackage[hyphens]{url}  
\usepackage{graphicx} 
\usepackage{natbib}  
\usepackage{caption} 
\usepackage{algorithm}
\usepackage{algorithmic}

\usepackage{booktabs}
\usepackage[table,xcdraw]{xcolor}
\usepackage{array}
\usepackage{multirow}
\usepackage{tabularx}

\usepackage{amssymb}
\usepackage{xspace}
\usepackage{longtable}
\usepackage{amsmath}
\usepackage[table,xcdraw]{xcolor}
\usepackage{subfigure}
\usepackage{subcaption}

\definecolor{myred}{HTML}{C00000}
\definecolor{myblue}{HTML}{002060}

\newcommand{\frameworkname}{\texttt{Med-PMC}\xspace}
\newcommand{\patientagent}{\textit{patient-actor} }

\usepackage{newfloat}
\usepackage{listings}
\DeclareCaptionStyle{ruled}{labelfont=normalfont,labelsep=colon,strut=off} 
\floatstyle{ruled}
\newfloat{listing}{tb}{lst}{}
\floatname{listing}{Listing}
\title{\frameworkname: Medical Personalized Multi-modal Consultation with a Proactive Ask-First-Observe-Next Paradigm}
\author{  Hongcheng Liu\equalcontrib\textsuperscript{\rm 1}, , Yusheng Liao\equalcontrib\textsuperscript{\rm 1,2}, Siqv Ou\textsuperscript{\rm 1}, Yuhao Wang\textsuperscript{\rm 1}, Heyang Liu\textsuperscript{\rm 1}\\ Yu Wang\textsuperscript{\rm 1,2}\thanks{Yu Wang is the corresponding author.}, Yanfeng Wang\textsuperscript{\rm 1,2}
}
\affiliations{  \textsuperscript{\rm 1} Shanghai Jiao Tong University,
\textsuperscript{\rm 2} Shanghai AI lab
}

\usepackage{bibentry}

\begin{document}

\maketitle

\begin{abstract}
The application of the Multi-modal Large Language Models~(MLLMs) in medical clinical scenarios remains underexplored. Previous benchmarks only focus on the capacity of the MLLMs in medical visual question-answering~(VQA) or report generation and fail to assess the performance of the MLLMs on complex clinical multi-modal tasks. In this paper, we propose a novel \textbf{Med}ical \textbf{P}ersonalized \textbf{M}ulti-modal \textbf{C}onsultation~(\frameworkname) paradigm to evaluate the clinical capacity of the MLLMs. \frameworkname builds a simulated clinical environment where the MLLMs are required to interact with a patient simulator to complete the multi-modal information-gathering and decision-making task. Specifically, the patient simulator is decorated with personalized actors to simulate diverse patients in real scenarios. We conduct extensive experiments to access 12 types of MLLMs, providing a comprehensive view of the MLLMs’ clinical performance. We found that current MLLMs fail to gather multimodal information and show potential bias in the decision-making task when consulted with the personalized patient simulators. Further analysis demonstrates the effectiveness of \frameworkname, showing the potential to guide the development of robust and reliable clinical MLLMs. Code and data are available at \url{https://github.com/LiuHC0428/Med-PMC}.

\end{abstract}

%

\section{Introduction}
The application of large language models (LLMs) in the medical field has been extensively explored. Models such as GPT-4~\citep{gpt4} and Med-Palm~\citep{singhal2022large} have been deployed for a wide range of clinical tasks including medical examinations~\citep{jin2021disease, pmlr-v174-pal22a}, medical consultation~\citep{liao2023automatic, tu2024towards}, and decision support~\citep{benary2023leveraging, hager2024evaluation}. Specifically, previous works utilize the role-playing capacity of the LLMs to work as a patient simulator~\cite{tu2024towards, liao2024automatic} and thus can build a simulated clinical environment to better evaluate the models' clinical performance. However, it is often necessary to handle a variety of multi-modal information in real clinical scenarios, such as medical images~(e.g. X-ray, CT) and biological signals~(e.g. temperature, ECG), to conduct a comprehensive analysis of patients. The limitation of LLMs in understanding only textual information restricts their application in broader clinical settings. 

Despite Multi-modal Large Language Models~(MLLMs) having capabilities in understanding multi-modal information, their application in medical clinical scenarios remains underexplored. Most of the existing medical multi-modal benchmarks only focus on gathering the information from medical images like visual question-answering~(VQA)~\citep{lau2018dataset, liu2021slake, he-etal-2021-towards} and report generation~\citep{thawkar2023xraygpt,hamamci2024ct2rep}. Other works handle relatively complex tasks, including reasoning utilizing the multi-modal information~\citep{li2024llava} or multiple images~\cite{wu2023towards}. However, these tasks are conducted in a static environment, which fails to simulate the interactions required with patients in real-world clinical settings. Consequently, these approaches fail to assess the performance and application potential of MLLMs in actual clinical scenarios. 


In this paper, we propose a novel \textbf{Med}ical \textbf{P}ersonalized \textbf{M}ulti-modal \textbf{C}onsultation~(\frameworkname) paradigm to build a simulated clinical environment and assess the performance of the MLLMs in clinical scenarios more comprehensively. Specifically, \frameworkname requires MLLMs to perform multi-turn decision reasoning based on the initial information provided by the patient. Within a limited number of interactions, the MLLMs must maximize the acquisition of the patient's multimodal symptom information and ultimately provide possible diagnostic results and treatment recommendations. To make the clinical environment modeled by \frameworkname more reliable, we propose a \patientagent agent. During interactions, the \patientagent first detects the type of action taken by the doctor and then generates relevant information from the case based on this action type. Basic information is extracted directly and responded to after being imbued with a unique identity and personality through the actor module. For examination and test information, the respective departments will directly provide reports. The \patientagent can not only record the doctor's decision-making process, ensuring the simulation's reliability, but also further simulates the diversity of patients in clinical scenarios. By incorporating \patientagent agents, \frameworkname can more accurately evaluate how MLLMs handle diverse and dynamic clinical interactions, ultimately improving the assessment of their real-world applicability and effectiveness in providing medical care. In conclusion, the contribution of this paper can be summarized as below:
\begin{itemize}
    \item \textbf{A Multi-modal Clinical Interactive Evaluation Framework}: We propose a novel evaluation framework, \textbf{Med}ical \textbf{P}ersonalized \textbf{M}ulti-modal \textbf{C}onsultation~(\frameworkname), which simulates a clinical environment to comprehensively assess the performance of MLLMs in real-world clinical scenarios. This framework addresses the limitations of existing approaches by incorporating multi-turn decision reasoning and multi-modal information.

    \item \textbf{Development of the reliable \patientagent Agent}: To enhance the reliability of the simulated clinical environment, we introduce the \patientagent agent. This agent dynamically interacts with the MLLMs, detecting the type of action taken by the doctor and generating relevant information based on the case. The \patientagent agent also simulates patient diversity, providing a more realistic and varied clinical setting for evaluating MLLMs.

    \item \textbf{Comprehensive Assessment of MLLMs' Clinical Performance}: By incorporating multi-turn reasoning, multimodal information, and patient diversity, \frameworkname allows for a more comprehensive assessment of MLLMs' capabilities in handling complex and dynamic clinical interactions. The extensive experiments assessing 12 types of MLLMs provide a holistic view of the MLLMs' clinical performance. Our results demonstrate the effectiveness of \frameworkname, thereby guiding the development of more robust and reliable MLLMs for clinical use.
\end{itemize}
\section{Related Works}
In the realm of medical language model development, significant strides have been made
by prior works such as HuatuoGPT~\citep{zhang2023huatuogpt} and Disc-medllm~\citep{bao2023disc}. HuatuoGPT employs a method that uses multi-turn consultation conversations generated by ChatGPT-like models for training data, whereas Disc-medllm simplifies the dialogue process into three distinct phases: information inquiry, preliminary diagnosis, and treatment suggestion. Similarly, other studies have proposed interaction frameworks to assess the clinical consultation performance of models by leveraging the role-playing capabilities of large language models (LLMs). \citet{johri2024craftmd} and \citet{liao2023automatic} have demonstrated the effectiveness of such frameworks in evaluating model performance in clinical settings. Furthermore, \citet{liao2024automatic} employed human evaluation to validate the rationality of the interaction framework and proposed a set of action categories for doctor models during the consultation process, enhancing the interpretability of model behavior. In addition, \citet{schmidgall2024agentclinic} considered the integration of multi-modal information during the consultation and diagnosis process. However, their approach does not fully evaluate the ability of multimodal large language models (MLLMs) to interpret medical images, which limits the comprehensiveness of their assessment.

\section{\frameworkname Evaluation Framework}
\begin{figure*}
    \centering
    \includegraphics[width=0.93\linewidth]{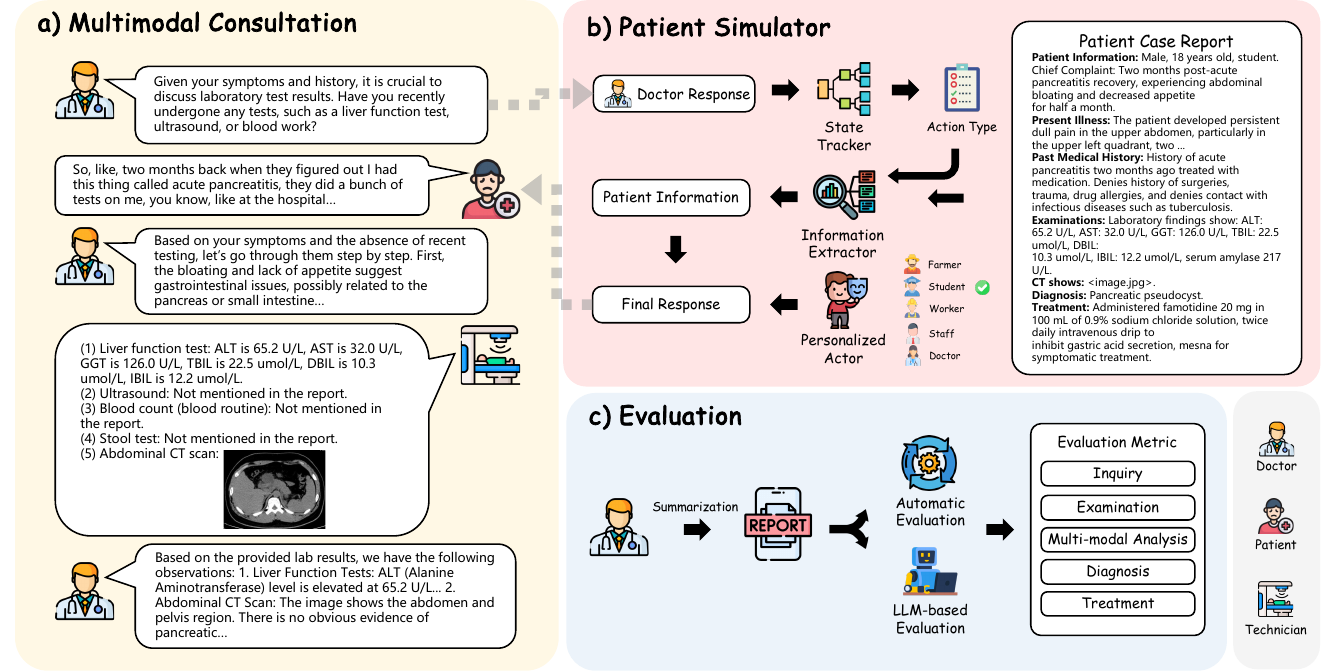}
    \caption{Overview of the \frameworkname evaluation framework. The whole framework can be divided into three parts, including a) Multi-modal consultation, b) Patient Simulator, and c) Evaluation.}
    \label{fig:pipeline}
\end{figure*}

To close the gap between the evaluation methods and actual clinical scenarios, we proposed \frameworkname evaluation framework to build a simulated clinical environment. As shown in Figure~\ref{fig:pipeline}, the MLLMs are required to interact with the personalized \patientagent agent and the multi-modal technician agent iteratively until it gathers enough information to make the final decision. Finally, MLLMs generate the diagnosis results and recommendations based on the gathered information.

\begin{table}[t]
\resizebox{\linewidth}{!}{%
\begin{tabular}{cc}
\hline
\multicolumn{1}{c}{State} & \multicolumn{1}{c}{Meanings}  \\ \hline
Begin     & The start of the consultation     \\
Effective Inquiry & Specific and relevant patient questions\\
Ineffective Inquiry & Specific but non-relevant patient questions\\
Ambiguous Inquiry   & Board patient questions           \\
Effective Advice & Specific and relevant examination questions\\
Ineffective Advice & Specific but non-relevant examination questions\\
Ambiguous Advice   & Board examination questions       \\
Other Topic     & Other questions      \\ \hline
\end{tabular}
}
\caption{The states definitions in the state detection stage.}
\label{tab:state_detection}
\end{table}

\subsection{Problem Formulation}
The proposed \frameworkname evaluation framework builds a simulated clinical environment upon $N$ realistic patient cases, which can be noted as $\mathcal{M}=\{\mathcal{I},\boldsymbol{d},\boldsymbol{t}\}^N$. $\mathcal{I}$ is the patient information, $\boldsymbol{d}$ is the diagnosis results, and $\boldsymbol{t}$ is the treatment recommendations. The patient information $\mathcal{I}$ contains three different types of multi-modal messages:
\begin{equation}
    \mathcal{I}=(\boldsymbol{i}_p, \boldsymbol{i}_e, \boldsymbol{i}_m)
\end{equation}
where $\boldsymbol{i}_p, \boldsymbol{i}_e, \boldsymbol{i}_m$ represent the basic patient information, examination results, and medical image, respectively. 

For each case, MLLMs are required to interact with the \patientagent agent for up to $M$ rounds to gather patient information. In the $k$-th round interaction, the MLLMs $\theta_d$ output the sequence $\boldsymbol{y}_{d,k}$ based on the interaction history $\mathcal{H}_{k}$:
\begin{equation}
    \mathcal{P}_{\theta_d}(\boldsymbol{y}_{d,k}|\mathcal{H}_{k}) = \prod^{|\boldsymbol{y}_{d,k}|}_{j=1}\mathcal{P}_{\theta_d}(y_{d,k}^j|y_{d,k}^{<j},\mathcal{H}_{k})
\end{equation}
and the \patientagent agent or technician $\theta_p$ responses according to the patient information in an autoregressive manner:
\begin{equation}
    \mathcal{P}_{\theta_p}(\boldsymbol{y}_{p,k}|\mathcal{I},\mathcal{H}_{k},\boldsymbol{y}_{d,k}) = \prod^{|\boldsymbol{y}_{p,k}|}_{j=1}\mathcal{P}_{\theta_p}(y_{p,k}^j|y_{p,k}^{<j},\mathcal{I},\mathcal{H}_{k},\boldsymbol{y}_{d,k})
\end{equation}
The interaction history is updated as follows:
\begin{equation}
    \mathcal{H}_{k} \cup (\boldsymbol{y}_{d,k}, \boldsymbol{y}_{p,k}) \rightarrow \mathcal{H}_{k+1}
\end{equation}
Once the doctor MLLMs have gathered sufficient patient information with $K$ turns consultation, they are required to generate a summarized report $\mathcal{R}$ and make the final decision on diagnosis and treatment:
\begin{equation}
    \mathcal{R} = \theta_d(\mathcal{H}_{K})
\end{equation}
We will evaluate the performance of the MLLMs on clinical tasks by comparing the content of their reports to the ground truth diagnosis results $\boldsymbol{d}$ and treatment plans $\boldsymbol{t}$.

\subsection{\textit{Patient-Actor} Agent}

To create a realistic and reliable clinical simulation environment, the \patientagent agent needs to exhibit appropriate behaviors and responses. In practice, patients generally engage in two types of behaviors: answering questions posed by the doctor, which typically involve providing basic personal information and describing symptoms, and completing specific examinations or tests as requested by the doctor. Therefore, the \patientagent agent should first classify the action of the doctor and then exhibit the expected behaviors, generating contextually relevant responses. Moreover, the patient agent also needs to have personalized characteristics, as clinical patients are diverse. Based on these requirements, we have designed the \patientagent agent to consist of three main components: a state tracker, a response generator, and a personalized actor.

\subsubsection{State Tracker \& State Definition}

As previously discussed, the state tracker categorizes doctors' actions to guide subsequent behaviors from three perspective. ($i$) The state tracker first determines whether the action is \textbf{Specific} or \textbf{Broad}. If the doctor's language is clear and specific, the patient-actor can respond to the doctor's action. On the contrary, if the language is vague and broad, the patient-actor should require the doctor to ask a more specific question. ($ii$) Following the real-world scenarios, we classify the specific actions into two main categories: \textbf{Basic Information Inquiry} and \textbf{Medical Examination Recommendations}. This classification determines whether the patient agent directly responds to questions or undergoes the relevant examinations. ($iii$) Finally, we categorize actions based on whether they retrieve relevant patient information into \textbf{Relevant} and \textbf{Irrelevant} types. If the information or the suggested examination/test results requested by the doctor are present in the patient's information, the action is considered Relevant; otherwise, the action is Irrelevant. In summary, we have defined a total of eight types of doctor actions to enable the \patientagent agent to respond more reliably. The types of actions and their meanings are detailed in Table~\ref{tab:state_detection}.

\subsubsection{Response Generator}

The response generator can produce corresponding replies based on the state tracker's classification of the doctor's actions. During experiments, we observed that some doctor models tend to find shortcuts to gather patient information by repeatedly asking vague questions like `Is there any other information?'. For such a situation, the generator should respond with requiring more specific action. The response generator can directly respond with relevant patient information or obtain the examination results from the technician agent for Basic Information Inquiries and Medical Examination Recommendations, respectively. Finally, For Irrelevant actions that request unavailable information, the \patientagent agent should respond `I don't know' to avoid fabricating false information. 

\subsubsection{Personalized Actor}
After obtaining the response information from the response generator, the Personalized Actor will rewrite the response, imbuing the patient agent with specific tones and personalities to simulate the diversity in how different patients express themselves in clinical settings. Specifically, we have set up 10 personas by combining two genders with five distinct professions: farmer, student, worker, office worker, and doctor. This allows us to model patients of varying ages, educational levels, socioeconomic statuses, living environments, and professional backgrounds, thereby capturing a wide range of patient expressions and behaviors.

\subsubsection{Technician Agent}
The technician agent aims to answer the questions about the examination, such as blood tests and X-ray images. To provide a more appropriate result rather than the external result without asking, we divided this process into two steps, one for examination detection and another for result provision, which is shown in Figure~\ref{fig:pipeline}. The result provision step needs two agents to complete the initial response based on the detected state and the final response based on the initial response. Similarly, we only provide the examination results for the technician agent and it can be activated only when the detected state is advice.
\section{Experiments}
\subsection{Datasets} 
We collected 30 medical cases from the medical website\footnote{\url{https://www.iiyi.com/}}, primarily focusing on the Department of General Surgery. For each case, patient information is divided into seven parts: Chief Complaint, Present Illness, Medical History, Examination, Diagnosis, Treatment, and Image Report. The ages of the patients range from 15 to 81 years, with a gender ratio of 12 males to 18 females. Each case includes 1-2 radio medical images, including MRI, ultrasound, and X-ray. 

\subsection{Experimental Settings}
We evaluate 12 MLLMs as the doctor models, which can complete the multi-modal multi-turn dialogue. We adopt the Qwen-Max~\citep{qwen1.5} as the backbone for the \patientagent agent, and we use GPT-4o~\citep{GPT-4o} as the evaluation LLMs. In practice, we set the seed as 0 to avoid randomness. As the \patientagent agent consultation, we both conduct the consultation from scratch and evaluate these two through the standard patient response to avoid the influence of the expression of the actor patient in the evaluation. 
\subsection{Evaluation Metrics}
We primarily assess the consultation capabilities of MLLMs from two perspectives: information gathering and final decision-making. Information gathering evaluates whether MLLMs can collect sufficient patient information during the consultation process, while final decision-making assesses whether MLLMs can ultimately provide accurate diagnoses and recommendations. Both automatic evaluation and LLM-based evaluation methods are adopted to validate the consultation capability of MLLMs.

\subsubsection{Auto-Metric Evaluation}
\paragraph{Information Gathering Metrics}
We categorize patient information into three main types: basic information, examination information, and multimodal (medical imaging) information. Basic information includes Chief Complaint, Present Illness, and Medical History. We assess the information-gathering capability of doctor MLLMs by calculating the recall score of these three types of information in the final report generated during the consultation. 
In practice, we use ROUGE-1~\citep{lin2004rouge} to calculate it.

\paragraph{Decision-Making Metrics}
Decision-making primarily encompasses two aspects: diagnosis and treatment plan. Similar to Information Gathering Metrics, we evaluate the model's decision-making capability by calculating the recall rate of diagnostic and treatment plan information from the patient data. However, unlike Information Gathering, diagnosis and treatment plans require the model to make predictions based on the patient information collected during the consultation, thus placing a greater emphasis on the model's reasoning ability.

\subsubsection{LLM-Based Evaluation}

We use the different prompts to set up various standards for inquiry, examination, multi-modal analysis, diagnosis, and treatment. 
The prompt we used in LLM-based evaluation is inspired by assessment plans from Peking University and National Health Commission of China~\citep{ Diagnostics, Standardized, OSCE}, and the score is on a scale of 1-5.

\begin{table*}[t]
\renewcommand{\arraystretch}{1.07} 
\centering
\setlength{\tabcolsep}{1.25mm}
\begin{tabular}{l|cccc|ccc|cc}
\toprule
 & \multicolumn{4}{c|}{\textbf{Information Gathering}} & \multicolumn{3}{c|}{\textbf{Decision-Making}} &  &  \\
\multirow{-2}{*}{\textbf{Model}} & \textbf{Inquiry} & \textbf{Exam.} & \textbf{MMA.} & \textbf{Avg.} & \textbf{Diagn.} & \textbf{Treat.} & \textbf{Avg.} & \multirow{-2}{*}{\textbf{Lens}} & \multirow{-2}{*}{\textbf{Turns}} \\
\midrule
\multicolumn{10}{c}{\cellcolor[HTML]{F2F2F2}\textit{Direct Prompt}} \\
\midrule
InternVL-1.5~\citep{internvl-1.5} & 52.42 & 13.84 & 13.94 & 26.73 & 26.32 & 21.44 & 23.88 & 36.47 & 8.63 \\
Qwen-VL-Chat~\citep{Qwen-VL-chat} & 57.48 & 15.06 & 2.46 & 25.00 & 29.47 & 23.98 & 26.73 & 32.62 & 10.00 \\
Mini-InternVL-1.5~\citep{mini-internvl-1.5} & 50.43 & 10.29 & 10.48 & 23.73 & 23.01 & 18.97 & 20.99 & 41.79 & 8.00 \\
HuatuoGPT-Vision-7B~\citep{chen2024huatuogpt-vision} & 54.87 & 10.95 & 5.14 & 23.65 & 32.78 & 25.41 & 29.10 & 44.44 & 9.76 \\
HuatuoGPT-Vision-34B~\citep{chen2024huatuogpt-vision} & 59.34 & 24.52 & 16.22 & 33.36 & 40.54 & 27.29 & 33.92 & 49.89 & 8.70 \\
GLM-4V~\citep{glm2024chatglm} & 32.19 & 10.73 & \textbf{25.06} & 22.66 & 21.75 & 19.97 & 20.86 & 65.51 & 7.90 \\
Qwen-VL-Max~\citep{Qwen-VL-max} & 56.34 & 19.66 & 15.20 & 30.40 & 28. 61 & 27.08 & 27.08 & 35.56 & 9.33 \\
Gemini1.5-Flash~\citep{reid2024gemini} & 63.99 & 17.95 & 19.25 & 33.73 & 37.81 & 26.75 & 32.28 & 27.39 & 9.53 \\
Gemini1.5-Pro~\citep{reid2024gemini} & 68.19 & 26.74 & 11.35 & 35.43 & 46.21 & \textbf{32.95} & 39.58 & 25.10 & 9.70 \\
GPT-4V~\citep{gpt4v} & 68.28 & 25.50 & 20.59 & 38.12 & 44.17 & 32.19 & 38.18 & 48.46 & 9.70 \\
GPT-4o~\citep{GPT-4o} & \textbf{68.75} & \textbf{32.50} & 20.76 & \textbf{40.67} & 45.12 & 32.25 & 38.69 & 39.54 & 9.43 \\
GPT-4o-mini~\citep{Mini-GPT-4o} & 68.25 & 25.20 & 6.45 & 33.30 & \textbf{46.83} & 32.17 & \textbf{39.50} & 30.04 & 9.00 \\
\midrule
\multicolumn{10}{c}{\cellcolor[HTML]{F2F2F2}\textit{Zero-Shot COT}} \\
\midrule
GLM-4V~\citep{glm2024chatglm} & 39.44 & 18.31 & \textbf{18.21} & 25.32 & 23.02 & 15.87 & 19.45 & 63.89 & 7.60 \\
Qwen-VL-Max~\citep{Qwen-VL-max} & 53.68 & 18.99 & 14.63 & 29.10 & 25.56 & 23.91 & 24.74 & 31.15 & 8.90 \\
Gemini1.5-Pro~\citep{reid2024gemini} & 67.32 & 21.70 & 6.13 & 31.72 & 44.03 & 31.00 & 37.52 & 23.59 & 8.90 \\
GPT-4o~\citep{GPT-4o} & \textbf{67.70} & \textbf{34.44} & \textbf{18.21} & \textbf{40.12} & \textbf{45.14} & \textbf{32.16} & \textbf{38.65} & 43.92 & 9.16 \\
\midrule
\multicolumn{10}{c}{\cellcolor[HTML]{F2F2F2}\textit{One-Shot COT}} \\
\midrule
GLM-4V~\citep{glm2024chatglm} & 36.49 & 25.24 & \textbf{27.93} & 29.89 & 26.24 & 19.66 & 22.95 & 130.36 & 7.80 \\
Qwen-VL-Max~\citep{Qwen-VL-max} & 56.45 & 27.42 & 14.76 & 32.88 & 30.54 & 28.56 & 29.55 & 60.71 & 8.23 \\
Gemini1.5-Pro~\citep{reid2024gemini} & 54.11 & 29.52 & 11.08 & 31.57 & 44.39 & 28.62 & 36.51 & 63.98 & 7.80 \\
GPT-4o~\citep{GPT-4o} & \textbf{70.24} & \textbf{45.93} & 25.31 & \textbf{47.16} & \textbf{47.42} & \textbf{31.68} & \textbf{39.55} & 64.78 & 8.66 \\
\bottomrule
\end{tabular}%
\caption{The consultation performance with the \textbf{standard \patientagent agent}. We evaluate different doctor models with three types of inference modes, including \textit{Direct Prompt}, \textit{Zero-COT}, and \textit{One-shot COT}. We assess the model's performance from two perspectives. `MMA.' indicates Multi-modal Analysis, `Diagn.' indicates Diagnosis, and `Treat' indicates Treatment. The best results of each dimension are \textbf{Bold}.}
\label{tab: main_result}
\end{table*}
\subsection{Consultation with Standard Patient}

\subsubsection{Direct Prompt}
We evaluated the consultation performance of 12 MLLMs with standard patients with the result shown in Table~\ref{tab: main_result}. Among these results, the GPT-4o and GPT-4v achieve the best performance in both information-gathering and decision-making evaluations, demonstrating their potential for application in medical consultations. The capabilities of the models across different dimensions are not consistent. Specifically, GLM-4V excels in medical image interpretation but is weaker in gathering information from inquiry and examination. Conversely, GPT-4o-mini performs exceptionally well in consultations (nearly the best), but its ability to analyze medical images is quite poor. On the other hand, the models' decision-making capabilities are relatively consistent. If a model can provide a good diagnosis, it is also likely to offer an effective treatment plan. In summary, the strongest existing MLLMs can collect a substantial amount of basic patient information and some examination data. However, even the most advanced medical MLLM, HuatuoGPT-Vision-34B, shows significant shortcomings in handling multimodal medical information. This highlights a new direction for improving future MLLMs in medical applications.

\subsubsection{Performance with CoT}
To comprehensively assess the capacity of the MLLMs, we further adopt the zero-shot CoT and one-shot COT to enhance the models' consultation ability. For zero-shot CoT, we use the specific prompt `Let's think step by step' to encourage the model to think through each step before generating a response. For the one-shot COT, we select an outside medical case and use GPT-4o to generate a CoT consultation dialogue between the doctor and the patient based on the full information in the report. We use the example as the one-shot CoT prompt and add it to the original doctor prompt to make the doctor models complete one-shot CoT. The CoT results are shown in Table~\ref{tab: main_result}. It can be observed that the performance improvement with zero-shot CoT is inconsistent. While there is an improvement for the weaker model, GLM-4V, other models do not show significant gains. In contrast, one-shot CoT noticeably enhances model performance, particularly in the ability to gather examination information.

\begin{table*}[t]
\centering
\setlength{\tabcolsep}{1mm}
\begin{tabular}{l|cccc|ccc}
\toprule
 & \multicolumn{4}{c|}{\textbf{Information Gathering}} & \multicolumn{3}{c}{\textbf{Decision}} \\
\multirow{-2}{*}{\textbf{Model}} & \textbf{Inquiry} & \textbf{Examination} & \textbf{MMA} & \textbf{Average} & \textbf{Diagnosis} & \textbf{Treatment} & \textbf{Average} \\
\midrule
\multicolumn{8}{c}{\cellcolor[HTML]{F2F2F2}\textit{Direct Prompt}} \\
\midrule
GLM-4V & 30.36 (\textcolor{myred}{-1.83}) & 17.86 (\textcolor{myblue}{+7.13}) & 22.46 (\textcolor{myred}{-2.60}) & 23.56 (\textcolor{myblue}{+0.90}) & 19.23 (\textcolor{myred}{-2.52}) & 17.72 (\textcolor{myred}{-2.25}) & 18.48 (\textcolor{myred}{-2.38}) \\
Qwen-VL-Max & 47.75 (\textcolor{myred}{-8.59}) & 20.22 (\textcolor{myblue}{+0.56}) & 19.61 (\textcolor{myblue}{+4.41}) & 29.19 (\textcolor{myred}{-1.21}) & 28.93 (\textcolor{myblue}{+0.32}) & 22.13 (\textcolor{myred}{-4.95}) & 25.53 (\textcolor{myred}{-2.31}) \\
Gemini1.5-Pro & 61.04 (\textcolor{myred}{-7.15}) & 23.89 (\textcolor{myred}{-2.85}) & 8.89 (\textcolor{myred}{-2.46}) & 31.27 (\textcolor{myred}{-4.15}) & 37.66 (\textcolor{myred}{-8.55}) & 28.64 (\textcolor{myred}{-4.31}) & 33.15 (\textcolor{myred}{-6.43}) \\
GPT-4o & 61.33 (\textcolor{myred}{-7.42}) & 29.62 (\textcolor{myred}{-2.88}) & 17.29 (\textcolor{myred}{-3.47}) & 36.08 (\textcolor{myred}{-4.59}) & 44.32 (\textcolor{myred}{-0.80}) & 28.66 (\textcolor{myred}{-3.59}) & 36.49 (\textcolor{myred}{-2.20}) \\
\midrule
\multicolumn{8}{c}{\cellcolor[HTML]{F2F2F2}\textit{Zero-Shot COT}} \\
\midrule
GLM-4V & 28.92 (\textcolor{myred}{-10.52}) & 13.83 (\textcolor{myred}{-4.48}) & 19.28 (\textcolor{myblue}{+1.07}) & 20.68 (\textcolor{myred}{-4.64}) & 18.09 (\textcolor{myred}{-4.93}) & 18.17 (\textcolor{myblue}{+2.30}) & 18.13 (\textcolor{myred}{-1.31}) \\
Qwen-VL-Max & 46.24 (\textcolor{myred}{-7.44}) & 25.09 (\textcolor{myblue}{+6.10}) & 15.44 (\textcolor{myblue}{+0.81}) & 28.92 (\textcolor{myred}{-0.18}) & 27.27 (\textcolor{myblue}{+1.71}) & 21.75 (\textcolor{myred}{-2.16}) & 24.51 (\textcolor{myred}{-0.23}) \\
Gemini1.5-Pro & 62.50 (\textcolor{myred}{-4.82}) & 25.56 (\textcolor{myblue}{+3.86}) & 9.09 (\textcolor{myblue}{+2.96}) & 32.38 (\textcolor{myblue}{+0.67}) & 45.87 (\textcolor{myblue}{+1.84}) & 30.75 (\textcolor{myred}{-0.25}) & 38.31 (\textcolor{myblue}{+0.80}) \\
GPT-4o & 60.91 (\textcolor{myred}{-6.79}) & 35.18 (\textcolor{myblue}{+0.74}) & 9.01 (\textcolor{myred}{-9.20}) & 35.03 (\textcolor{myred}{-5.08}) & 44.44 (\textcolor{myred}{-0.70}) & 30.09 (\textcolor{myred}{-2.07}) & 37.27 (\textcolor{myred}{-1.38}) \\
\midrule
\multicolumn{8}{c}{\cellcolor[HTML]{F2F2F2}\textit{One-Shot COT}} \\
\midrule
GLM-4V & 31.56 (\textcolor{myred}{-4.93}) & 19.00 (\textcolor{myred}{-6.24}) & 30.50 (\textcolor{myblue}{+2.57}) & 27.02 (\textcolor{myred}{-2.87}) & 24.00 (\textcolor{myred}{-2.24}) & 18.86 (\textcolor{myred}{-0.80}) & 21.43 (\textcolor{myred}{-1.52}) \\
Qwen-VL-Max & 42.91 (\textcolor{myred}{-13.54}) & 28.27 (\textcolor{myblue}{+0.85}) & 9.93 (\textcolor{myred}{-4.83}) & 27.04 (\textcolor{myred}{-5.84}) & 22.36 (\textcolor{myred}{-8.18}) & 21.13 (\textcolor{myred}{-7.43}) & 21.75 (\textcolor{myred}{-7.81}) \\
Gemini1.5-Pro & 51.51 (\textcolor{myred}{-2.60}) & 21.52 (\textcolor{myred}{-8.00}) & 15.44 (\textcolor{myblue}{+4.36}) & 29.49 (\textcolor{myred}{-2.08}) & 32.49 (\textcolor{myred}{-11.90}) & 29.38 (\textcolor{myblue}{+0.76}) & 30.94 (\textcolor{myred}{-5.57}) \\
GPT-4o & 60.65 (\textcolor{myred}{-9.59}) & 33.05 (\textcolor{myred}{-12.88}) & 27.97 (\textcolor{myblue}{+2.66}) & 40.56 (\textcolor{myred}{-6.60}) & 49.57 (\textcolor{myblue}{+2.15}) & 31.60 (\textcolor{myred}{-0.08}) & 40.59 (\textcolor{myblue}{+1.04}) \\
\bottomrule
\end{tabular}%
\caption{The consultation performance with the \textbf{\patientagent agent}. We evaluate different doctor models with three types of inference modes, including \textit{Direct Prompt}, \textit{Zero-COT}, and \textit{One-shot COT}. The results in `$(\cdot)$' represent the delta performance compared to the standard patient-actor agent shown in Table 2. `MMA.' indicates Multi-modal Analysis.}
\label{tab: actor_results}
\end{table*}

\subsection{Consultation with \patientagent agent}
For the consultation with \patientagent agent, we select 4 categories of MLLMs with 3 inference methods to validate the impact of the personalized patient in the consultation process. As shown in Table~\ref{tab: actor_results}, under the influence of a personalized actor, the performance of most MLLMs and inference methods declined. In terms of information gathering, the collection of inquiries suffered the most significant drop, as the information obtained through questioning is directly affected by the patient's expression. Consequently, the decision-making performance also suffered significantly. 

A surprising finding is that Zero-shot CoT significantly mitigates the impact of the patient-actor, with decision-making performance declining by only -1.38 points, much less than Direct Prompt (-6.43) and One-Shot CoT (-7.81). This may be because the model's step-by-step reasoning before making a decision allows it to better handle information expressed in various ways. On the other hand, the failure of One-Shot CoT is likely due to the given example not including similar situations.

\begin{figure}[t]
    \centering
    \includegraphics[width = \linewidth]{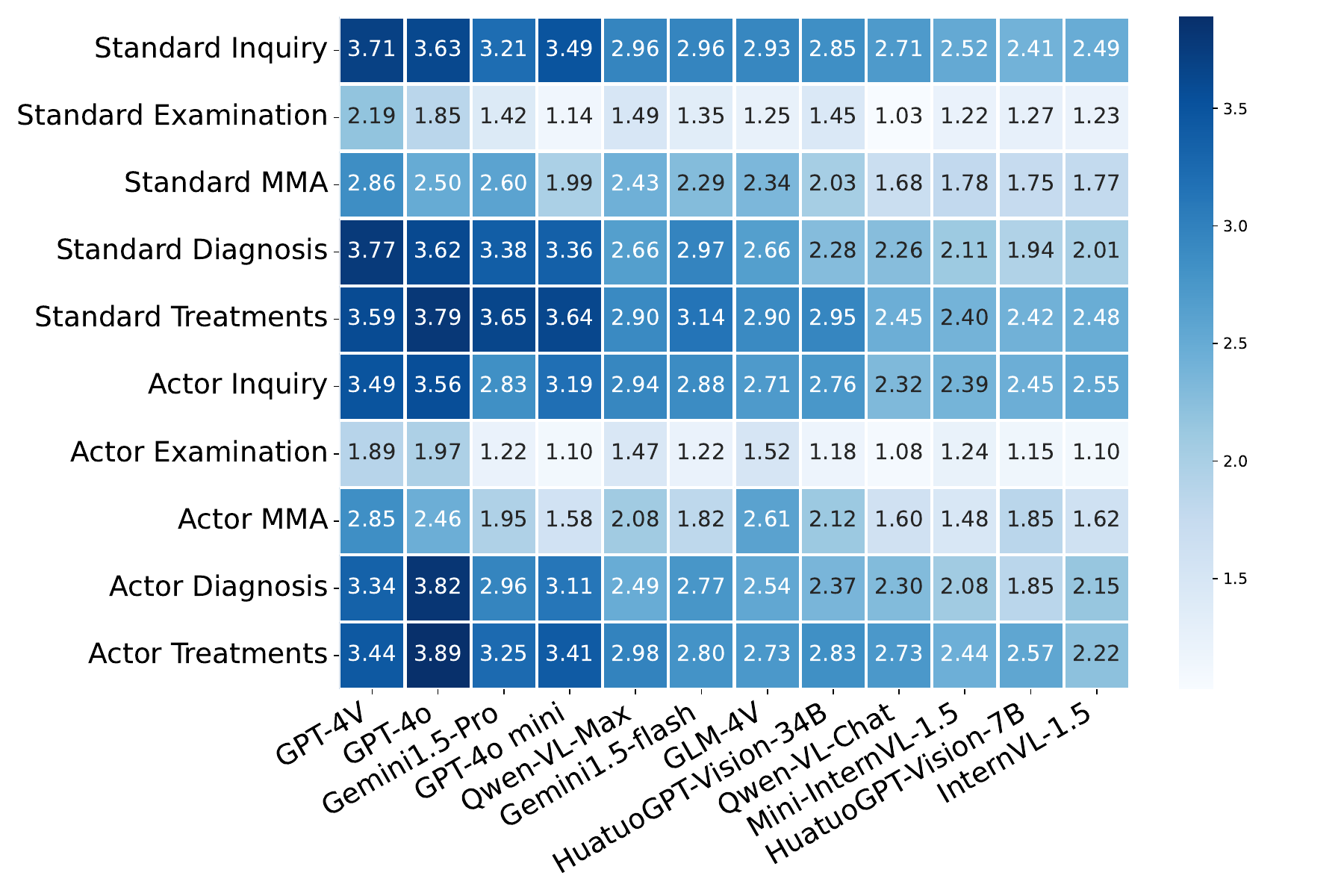}
    \caption{Results of LLM-based evaluation on consultation with both standard patient and actor patient.}
    \label{fig: llm evaluation}
\end{figure}

More LLM-based evaluation results of the standard patient and actor patient are shown in Figure~\ref{fig: llm evaluation}. The performance of the MLLMs consulting with standard patients is better than that of the patient actor, which is consistent with the automatic metrics. Besides, it is obvious that the MLLMs fall short of gathering the examination information. 

\subsection{Patient Evaluation}
To demonstrate the effectiveness of our evaluation framework, We evaluated the performance of the standard patient agent and actor agent in response generation, and the result is shown in Table~\ref{tab:patient-eval} and Table~\ref{tab:actor-eval}. In detail, we randomly select 3 different dialogues in each consultation case and use GPT-4o to evaluate the performance. As the standard patient, we consider five aspects: description, information completeness, cooperation, understanding, and communication ability. We assess the semantic consistency and character features between the standard and the actor output. The results in the two tables show that the patient agent can understand the doctor's purpose and generate an appropriate response, making the framework evaluate the doctor effectively.

\begin{table}[t]
\centering
\begin{tabular}{cccccc}
\toprule
D & I & C & U & CM & Total \\ \midrule
2.00 & 1.85 & 1.96& 1.81  & 1.85  & 9.48  \\ \bottomrule
\end{tabular}
\caption{The LLM-based evaluation result of standard patient agent performance. Each aspect is worth 2 points. D: Description clarity, I: Information completeness, C: Cooperation, U: Understanding, CM: communication ability}
\label{tab:patient-eval}
\end{table}

\begin{table}[t]
\centering
\begin{tabular}{ccc}
\toprule
Semantic Consistency & Character Feature & Total \\ \midrule
4.85   & 4.59  & 9.44  \\ \bottomrule
\end{tabular}
\caption{The LLM-based evaluation result of actor agent performance. Each aspect is worth 5 points.}
\label{tab:actor-eval}
\end{table}

\begin{figure*}[t]
    \centering
    \subfigure[]{
    \includegraphics[width = 0.4\linewidth]{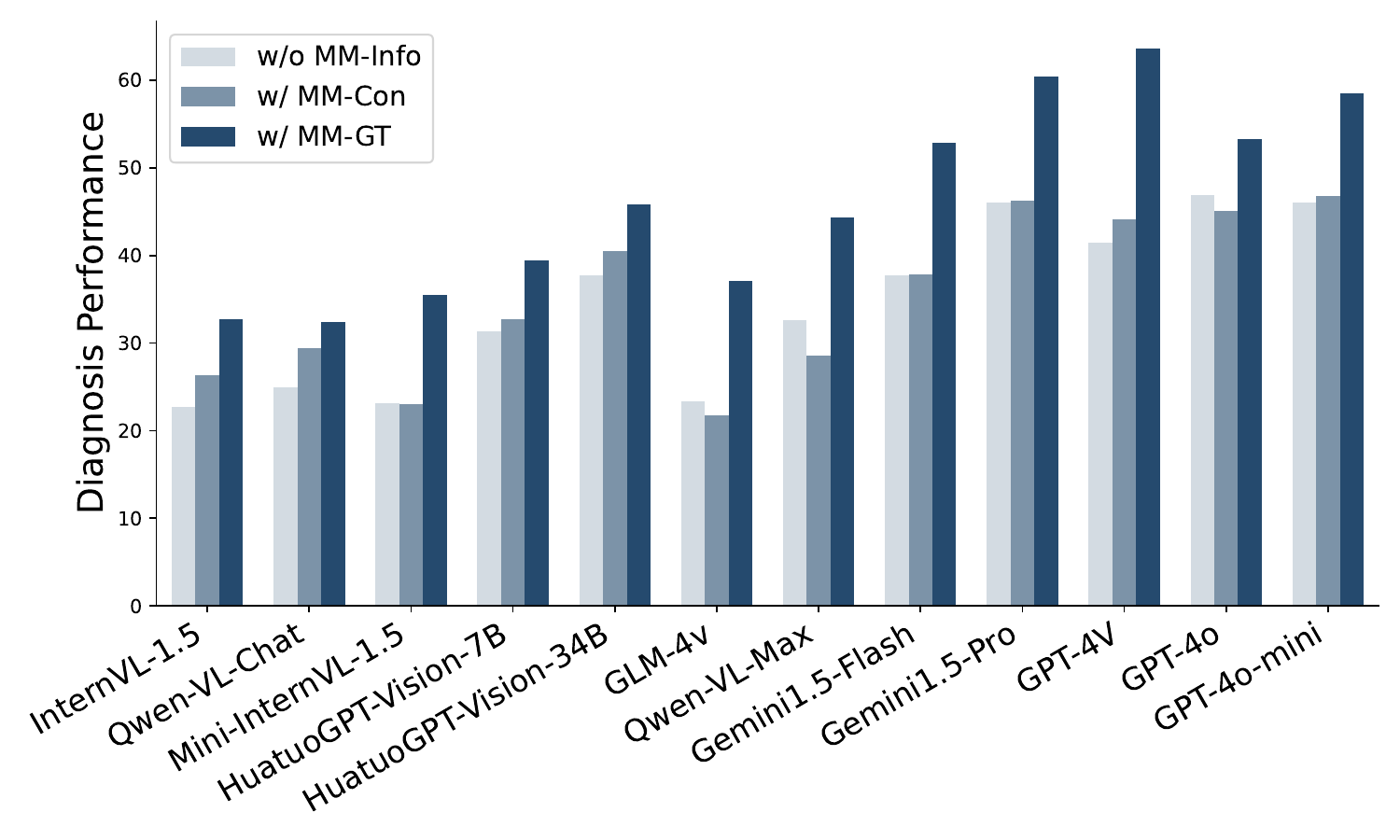}}
    \quad
    \subfigure[]{
    \includegraphics[width = 0.4\linewidth]{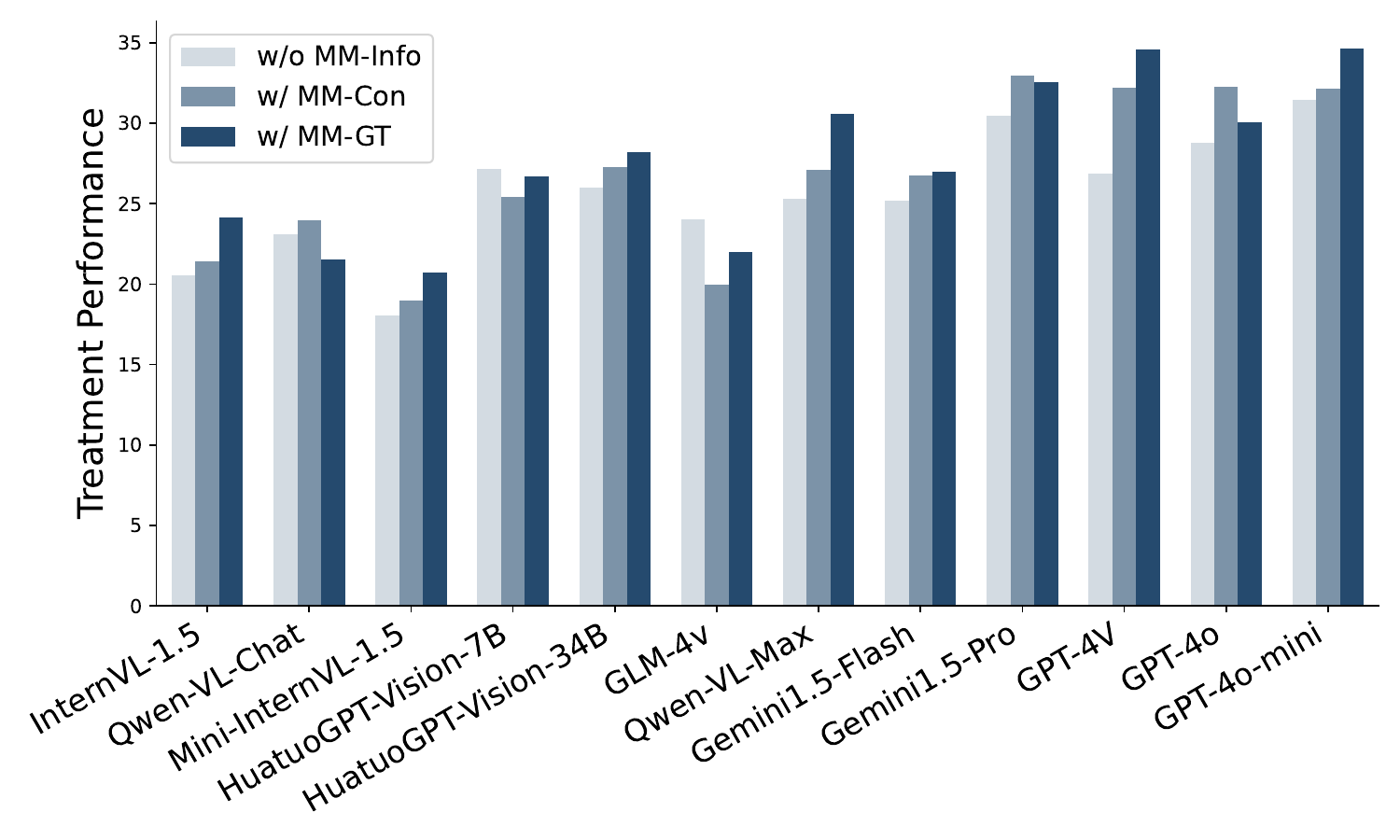}}
    \quad
    \caption{Ablation study of multi-modal information on (a) Diagnosis performance and (b) Treatment performance. We compare three situations to show the impact of the multi-modal information on the consultation results. Specifically, `w/o MM-Info' represents the model without any multi-modal information, `w/MM-Con' denotes the model with multi-modal information obtained through consultation, and `w/MM-GT' signifies the model with complete multi-modal information.}
    \label{fig: mm_ablation}
\end{figure*}

\section{Discussion}
\subsection{Gender Bias of the MLLMs}
\begin{figure}[t]
    \centering
    \includegraphics[width = \linewidth]{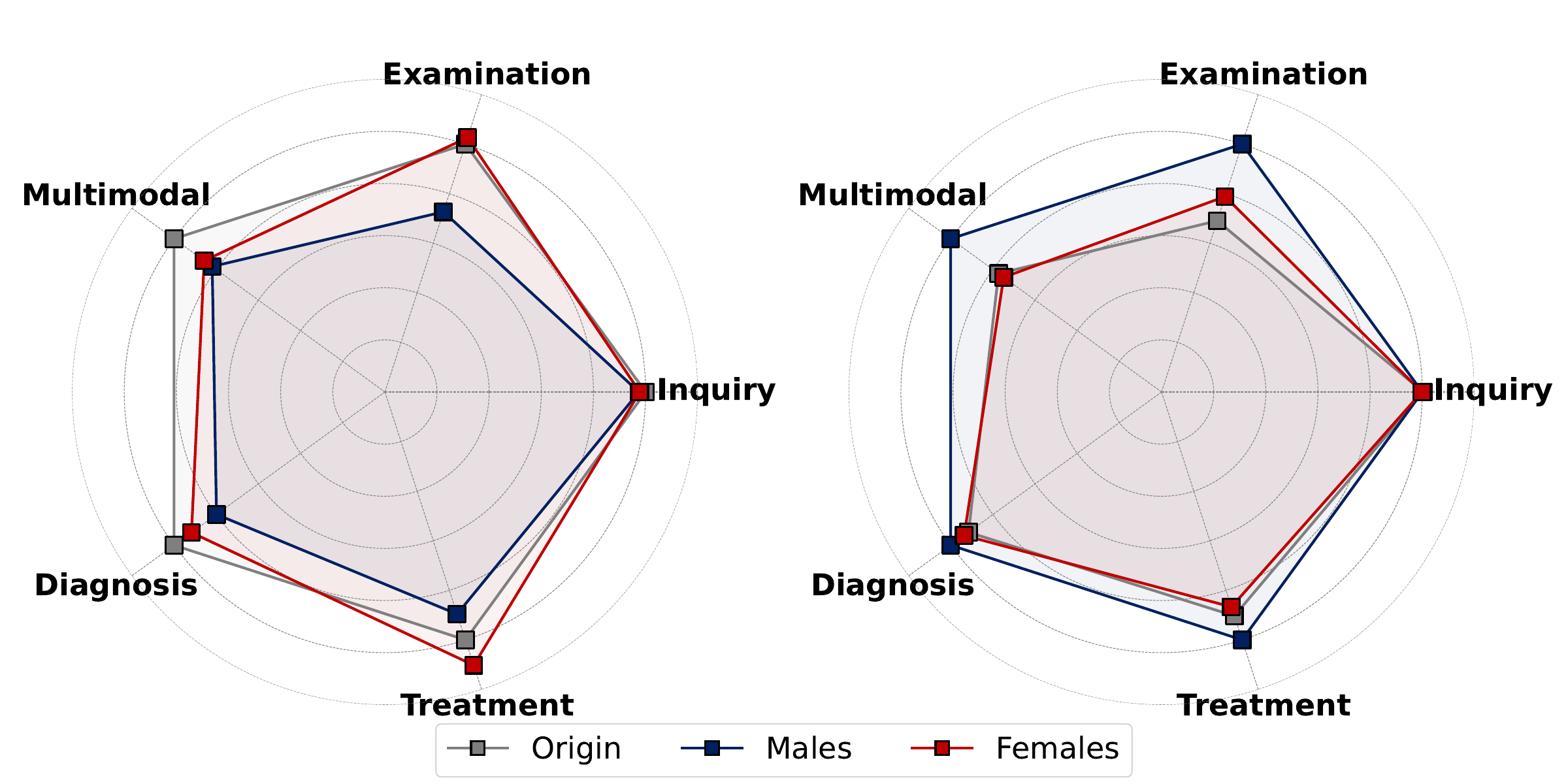}
    \caption{Gender bias of MLLMs. On the left is the GPT-4o. On the right is the Gemini1.5-Pro. Both MLLMs exhibit varying degrees of gender bias in medical consultations. All the scores are normalized.}
    \label{fig: gender_bias}
\end{figure}
After analysis the impact of the personalized actor, we further explore the gender bias on Gemini-pro and GPT-4o in medical consultation. Specifically, we set the gender of all patients in the entire test set to either male or female to observe any potential gender bias in the model.\footnote{As these diseases are not sensitive to gender, this setting about gender cannot influence the final result.} The results in Figure~\ref{fig: gender_bias} show that both models exhibit varying degrees of gender bias. Specifically, GPT-4o performs worse with male patients, while Gemini1.5-Pr performs worse with female patients. Notably, gender has almost no impact on the inquiry performance. As a result, the gender effect can significantly influence model performance in practice. Therefore, addressing and mitigating gender bias should be a key focus for future research and development efforts.

\subsection{Effectiveness of Multi-modal Information}
In this section, we primarily analyze the importance of multimodal information in decision-making. We set up three scenarios: 1) \textbf{\textit{w/o} MM-Info}: MLLMs have no access to the patient's multimodal information. 2) \textbf{\textit{w/} MM-Con}: MLLMs can acquire and analyze the patient's multimodal information through consultation, representing the standard scenario. 3) \textbf{\textit{w/} MM-GT}: MLLMs have direct access to the patient's medical imaging reports.
By comparing the performance differences of MLLMs in decision-making across these three scenarios, we can evaluate the effectiveness of multimodal information. The results are shown in Table~\ref{fig: mm_ablation}. On one hand, multimodal information significantly impacts diagnosis but has a lesser effect on treatment. The `w/ MM-GT' scenario notably enhances MLLMs' decision-making performance. On the other hand, the `w/ MM-Con' scenario often performs worse than the `w/o MM-Info' scenario. This indicates that MLLMs are currently unable to effectively analyze medical images, falling short of clinical standards. In summary, while multimodal information is crucial for clinical diagnosis, current MLLMs cannot fully utilize medical imaging information to make accurate diagnoses.

\begin{figure}
    \centering
    \includegraphics[width = .85\linewidth]{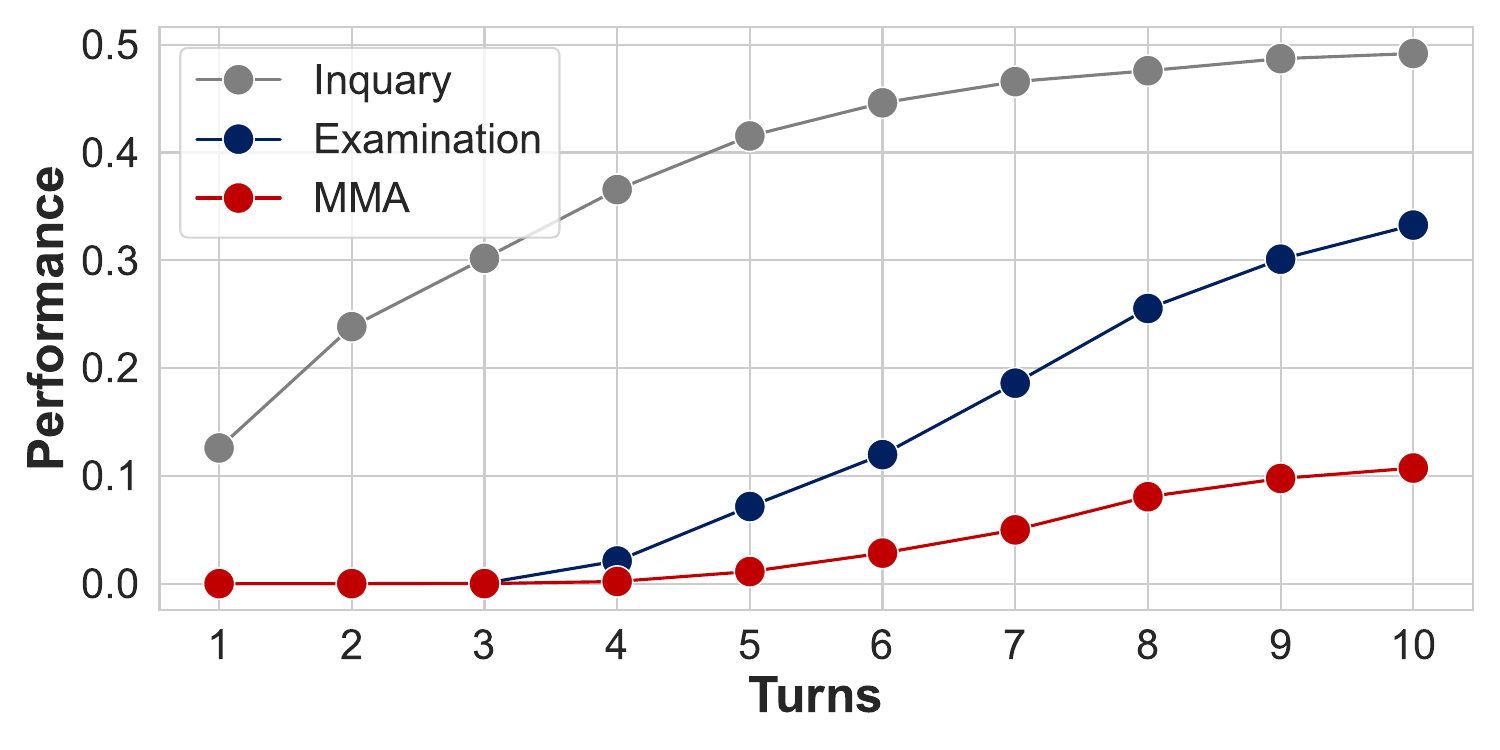}
    \caption{Information gathering performance changes with the consultation turns. All the scores are the averaged results of 12 types of MLLMs.}
    \label{fig: information gain}
\end{figure}

\subsection{Information Gains with Consultation Progress}
We further explored how models collect different types of information during the consultation process. As shown in Figure~\ref{fig: information gain}, we compiled the cumulative information collected by all models at each round and found that MLLMs tend to gather general information first, followed by examination information, and finally image information. This indicates that MLLMs possess a basic logical structure for consultations, and their preferences in clinical tasks are similar to those of human doctors.

\section{Conclusions}
In this paper, we proposed a Multi-modal Consultation Evaluation Framework to assess the clinical performance of the MLLMs and explore their potential application in realistic clinical environments. The results show that current MLLMs fail shot in consulting with personalized actor and can not fully utilize the multi-modal information. Our results demonstrate the effectiveness of \frameworkname , thereby guiding the development of more robust and reliable MLLMs for clinical use.

\section{Ethical Statement}
The medical cases are collected from the iiyi website, where doctors voluntarily upload and share information. This data is explicitly permitted for use in research and educational purposes. To ensure the protection of patient privacy, our dataset does not include any personally identifiable information such as patient names, hospital details, or any other sensitive information. Consequently, there is no risk of privacy breaches associated with our dataset. Additionally, all data usage complies with ethical standards and regulations governing medical information and research.

\bigskip

\bibliography{aaai25}

\clearpage
\section{Reproducibility Checklist}
This paper
\begin{itemize}
    \item Includes a conceptual outline and/or pseudocode description of AI methods introduced. \textcolor{red}{yes}
    \item Clearly delineates statements that are opinions, hypothesis, and speculation from objective facts and results. \textcolor{red}{yes}
    \item Provides well marked pedagogical references for less-familiare readers to gain background necessary to replicate the paper. \textcolor{red}{yes}
\end{itemize}
Does this paper make theoretical contributions? \textcolor{red}{yes}\\
If yes, please complete the list below.
\begin{itemize}
    \item All assumptions and restrictions are stated clearly and formally. \textcolor{red}{yes}
    \item All novel claims are stated formally (e.g., in theorem statements). \textcolor{red}{yes}
    \item Proofs of all novel claims are included. \textcolor{red}{yes}
    \item Proof sketches or intuitions are given for complex and/or novel results. \textcolor{red}{yes}
    \item Appropriate citations to theoretical tools used are given. \textcolor{red}{yes}
    \item All theoretical claims are demonstrated empirically to hold. \textcolor{red}{yes}
    \item All experimental code used to eliminate or disprove claims is included. \textcolor{red}{yes}
\end{itemize}
Does this paper rely on one or more datasets? \textcolor{red}{yes}\\
If yes, please complete the list below.
\begin{itemize}
    \item A motivation is given for why the experiments are conducted on the selected datasets. \textcolor{red}{yes}
    \item All novel datasets introduced in this paper are included in a data appendix. \textcolor{red}{yes}
    \item All novel datasets introduced in this paper will be made publicly available upon publication of the paper with a license that allows free usage for research purposes. \textcolor{red}{yes}
    \item All datasets drawn from the existing literature (potentially including authors’ own previously published work) are accompanied by appropriate citations. \textcolor{red}{yes}
    \item All datasets drawn from the existing literature (potentially including authors’ own previously published work) are publicly available. \textcolor{red}{yes}
    \item All datasets that are not publicly available are described in detail, with explanation why publicly available alternatives are not scientifically satisficing. \textcolor{red}{NA.}
\end{itemize}
Does this paper include computational experiments? \textcolor{red}{yes}\\
If yes, please complete the list below.

\begin{itemize}
    \item Any code required for pre-processing data is included in the appendix. \textcolor{red}{yes}
    \item All source code required for conducting and analyzing the experiments is included in a code appendix. \textcolor{red}{yes}
    \item All source code required for conducting and analyzing the experiments will be made publicly available upon publication of the paper with a license that allows free usage for research purposes. \textcolor{red}{yes}
    \item All source code implementing new methods have comments detailing the implementation, with references to the paper where each step comes from \textcolor{red}{yes}
    \item If an algorithm depends on randomness, then the method used for setting seeds is described in a way sufficient to allow replication of results. \textcolor{red}{yes}
    \item This paper specifies the computing infrastructure used for running experiments (hardware and software), including GPU/CPU models; amount of memory; operating system; names and versions of relevant software libraries and frameworks. \textcolor{red}{yes}
    \item This paper formally describes evaluation metrics used and explains the motivation for choosing these metrics. \textcolor{red}{yes}
    \item This paper states the number of algorithm runs used to compute each reported result. \textcolor{red}{yes}
    \item Analysis of experiments goes beyond single-dimensional summaries of performance (e.g., average; median) to include measures of variation, confidence, or other distributional information. \textcolor{red}{yes}
    \item The significance of any improvement or decrease in performance is judged using appropriate statistical tests (e.g., Wilcoxon signed-rank). \textcolor{red}{no}
    \item This paper lists all final (hyper-)parameters used for each model/algorithm in the paper’s experiments. \textcolor{red}{yes}
    \item This paper states the number and range of values tried per (hyper-) parameter during development of the paper, along with the criterion used for selecting the final parameter setting. \textcolor{red}{NA}
\end{itemize}

\clearpage
\appendix

\section{The details on Doctor models}
\begin{table}[h]
\centering
\small
\renewcommand{\arraystretch}{1.15}
\begin{tabular}{lccc}
\hline
\multicolumn{1}{c}{Model} & Size  & Precision                   & Belongs   \\ \hline
Mini-InternVL-1.5         & 4.2B  & {\color[HTML]{4F4F4F} BF16} & OpenGVLab \\
InternVL-1.5              & 25.5B & {\color[HTML]{4F4F4F} BF16} & OpenGVLab \\
Gemini-flash              & API   & -                           & Google    \\
Gemini-pro                & API   & -                           & Google    \\
GLM4v                     & API   & -                           & THUDM     \\
GPT-4o                    & API   & -                           & OpenAI    \\
GPT-4o mini                 & API   & -                           & OpenAI    \\
GPT-4v                    & API   & -                           & OpenAI    \\
Qwen-vl-chat-v1           & 9.6B  & {\color[HTML]{4F4F4F} FP16} & Ali       \\
Qwen-vl-max               & API   & -                           & Ali     \\  
HuatuoGPT-Vision-7B  & 7.94B    & BF16                           & SRIBD    \\
HuatuoGPT-Vision-34B  & 34.8B   & BF16                           & SRIBD    \\  \hline
\end{tabular}
\caption{The details about the doctor models.}
\label{tab:llm_detail}
\end{table}
More details on doctor models are shown in Table~\ref{tab:llm_detail}. Furthermore, we set the temperature as 0.01 and ‘do\_sample=False’ to demonstrate the most primitive capabilities of the model, and we set the seed as 0 to make the results reproducible.

\section{The Details on Evaluation}
\subsection{Auto-Metric Formulation}
\begin{table}[t]

\renewcommand{\arraystretch}{1.15}
\centering
\begin{tabularx}{0.48\textwidth}{>{\hsize=0.15\hsize \centering}X>{\hsize=0.85\hsize}X}
\hline 
\textbf{Symbol} & \textbf{Significance} \\
\hline
$N$  & The total number of cases. \\
$K$  & The total turns of consultation. \\
\multirow{3}{*}{$\boldsymbol{i}_p$}     &   The ground truth of the inquiry, including the patient information, chief, present illness, and past medical history.           \\
$\boldsymbol{i}_e$ & The ground truth of the examination. \\
$\boldsymbol{i}_m$ & The ground truth of the image analysis. \\
$\boldsymbol{d}$ & The ground truth of the diagnosis. \\
$\boldsymbol{t}$ & The ground truth of the treatment. \\
 $\mathcal{R}$      &      The final report generated by the doctor agent.       \\
   $\boldsymbol{y}_{d}$ & The output generated by the doctor agent.\\
  $\boldsymbol{y}_{p}$     &    The response of the patient-actor agent.\\
  $\boldsymbol{y}_{t}$     &    The response of the technician agent. \\
  $\mathcal{E}$     &    The exam items.          \\
  $\rm {I_{i}}$    &    The amount of information gained in inquiry  \\ 
  $\rm {I_{e}}$    &    The amount of information gained in examination  \\
  $\rm {I_{m}}$    &    The amount of information gained in multi-modal analysis  \\\hline         

\end{tabularx}
\caption{The significance of different symbols.}
\label{tab:metric_sysbolic}
\end{table}

For the sake of convenience, we summarize the symbols used in the formula in Table~\ref{tab:metric_sysbolic}.
\subsubsection{Information Gathering}
\paragraph{Inquiry}
The inquiry measures how much patient information is obtained in the consultation and summarizes it in the report. 
\begin{equation}
    {\rm Iquiry}  =  \frac{1}{N} \sum_{j=1}^{N} ({\rm Recall}(\boldsymbol{i}_p^j,\mathcal{R}^j)),
\end{equation}
Where $\boldsymbol{i}_p^i$ denotes the ground truth of the inquiry, including the patient information, chief, present illness, and past medical history, $\mathcal{R}^j$ is the final report generated by the doctor agent, and the $N$ denotes the total number of cases. We use ROUGE-1~\citep{lin2004rouge} as the implementation of the recall calculation without special instructions.

\paragraph{Examination}
The examination measures how many examination items are made in the consultation and summarizes them in the report. We first detect the exam entity in the patient information, which is formulated as $e$. 
\begin{equation}
    \mathcal{E} = \{e | e \in {\rm NER}(\boldsymbol{i}_e)\},
\end{equation}   
where $\boldsymbol{g}_e$ denotes the ground truth of the examinations and $\rm NER$ denotes the entity extraction tool \texttt{Spacy}~\citep{Honnibal_spaCy_Industrial-strength_Natural_2020}. Then, we calculate the examination score:
\begin{equation}
    {\rm Examination} = \frac{1}{N} \sum_{j = 1}^{N}\frac{ \sum_{e\in\mathcal{E}^j} \mathbb{I} (e \in \mathcal{R}^j)}{|\mathcal{E}^j|},
\end{equation}
Where $\mathbb{I}(\cdot)$ equals 1 only when the equation in $(\cdot)$ is held and $|\mathcal{E}^j|$ denotes the number of entities in $\mathcal{E}^j$.
\paragraph{Multi-modal Analysis}
This metric measures how much the correlation between the doctor's response and the image report after obtaining the image.
\begin{equation}
    {\rm MMA} = \frac{1}{N} \sum_{j=1}^{N} ({\rm Recall}(\boldsymbol{i}_m^j,\boldsymbol{y}_{d,k}^{j})),
\end{equation}
where the $\boldsymbol{y}_{d,k}^{i}$ denotes the output after the doctor obtains the image in the $i^{th}$ case, i.e. the $k$ turns.
\subsubsection{Decision-Making}
\paragraph{Diagnosis}
The diagnosis measures the degree of diagnosis of the disease, such as the location of the disease and the type of disease.
\begin{equation}
    {\rm Diagnosis} = \frac{1}{N} \sum_{j=1}^{N} {\rm Recall_L}(\boldsymbol{d}^j,\mathcal{R}^j),
\end{equation}
where we use the ROUGE-L as the measure in recall calculation.
\paragraph{Treatment }
The treatment measures how much the treatment plan coincides with the ground truth.
\begin{equation}
    {\rm Treatment} =  \frac{1}{N} \sum_{j=1}^{N} {\rm Recall}(\boldsymbol{t}^j,\mathcal{R}^j).
\end{equation}
\subsection{LLM-based Prompts}
\begin{table*}[t]
\renewcommand{\arraystretch}{1.15}
\centering
\large
\resizebox{16cm}{!}{
\begin{tabularx}{\textwidth}{X}
\rowcolor[HTML]{343434} 
{\color[HTML]{FFFDFA} Promt for LLM-based Evaluation on Information Gathering}   
\\
\textbf{System prompt} \\
You are a strict evaluator. Below, a case will be presented, which has been divided into three modules: case info, dialogue, and case by doctor. Your task is to evaluate the consultation, examination, diagnosis and treatment, and communication skills based on the following evaluation criteria. Provide ratings for each aspect on a scale of 1 to 5 at the beginning in format: 'score: x'. In this case, the case info is the standard case content, which you should consider as a reference answer and use it as a benchmark to evaluate the conversation with the case summarized by the doctor. For cases where the criteria for corresponding scores are not fully met, one can take the decimal between them. You don't need to give any explanation or repeat my input. Use the following scale to evaluate each criterion:\\
\textbf{Inquiry}\\
score 1: The doctor's logic is confused, questions are unreasonable, and language is obscure with excessive medical jargon. Medical history collection is incomplete and lacks details.\\
score 2: The doctor has some logic but still makes unreasonable arrangements and uses obscure language with medical terms. Medical history collection is incomplete with missing key points.\\
score 3: The doctor has some logic with fewer unreasonable arrangements, speaks more fluently with less jargon, and collects most key medical history details.\\
score 4: The doctor is logical and clear, focuses on key issues, avoids jargon, and collects a complete medical history.\\
score 5: The doctor is very clear and logical, focuses on key issues, avoids jargon entirely, and collects a fully complete medical history.\\
\textbf{Examination}\\
score 1: The doctor performs no valid tests (i.e., the tests in the 'Examination' item in the standard case) and prescribes highly unnecessary tests.\\
score 2: The doctor performed the tests in the 'Examination' item in the standard case but some additional tests are unnecessary, with significantly inaccurate interpretations.\\
score 3: The doctor performed the tests in the 'Examination' item in the standard case but with a few unnecessary ones, and has somewhat inaccurate interpretations.\\
score 4: The doctor performed the tests in the 'Examination' item in the standard case but very few unnecessary ones, and slightly inaccurate interpretations.\\
score 5: The doctor performs all necessary tests with reasonable prescriptions and virtually error-free interpretations.\\
Attention: if doctor doesn't perform the examination in the 'Examination' item in the standard case, the score should be 1.\\
\textbf{Multi-modal Analysis}: \\
score 1: Compared to examination results in benchmarks, there is a very large gap between the doctor's analysis of the examination image, e.g., the conclusions given are very different, etc.\\
score 2: Compared to examination results in benchmarks, there is a large gap between the doctor's analysis of the examination image.\\
score 3: Compared to examination results in benchmarks, there are some diffrences between the doctor's analysis of the examination image, but the conclusions are similar.\\
score 4; Compared to examination results in benchmarks, there are small gaps between the doctor's analysis of the examination image.\\
score 5: Compared to examination results in benchmarks, doctor's analysis of the examination image were almost identical\\

\end{tabularx}}
\caption{The prompt for LLM-based evaluation on information gathering.}
\label{tab:llm_eval1_prompt}
\end{table*}

\begin{table*}[h]
\renewcommand{\arraystretch}{1.15}
\centering
\large
\resizebox{16cm}{!}{
\begin{tabularx}{\textwidth}{X}
\rowcolor[HTML]{343434} 
{\color[HTML]{FFFDFA} Promt for LLM-based Evaluation on Decision Making}   
\\
\textbf{System prompt} \\
You are a strict evaluator. Below, a case will be presented, which has been divided into three modules: case info, dialogue, and case by doctor. Your task is to evaluate the consultation, examination, diagnosis and treatment, and communication skills based on the following evaluation criteria. Provide ratings for each aspect on a scale of 1 to 5 at the beginning in format: 'score: x'. In this case, the case info is the standard case content, which you should consider as a reference answer and use it as a benchmark to evaluate the conversation with the case summarized by the doctor. For cases where the criteria for corresponding scores are not fully met, one can take the decimal between them. You don't need to give any explanation or repeat my input. Use the following scale to evaluate each criterion:\\
\textbf{Diagnosis}\\
Consider item 'Diagnosis' in the 'case info' as the standard diagnosis answer\\
score 1: The doctor is unable to provide a correct primary diagnosis and differential diagnosis, and it's far from the answer. e.g., pancreatitis is diagnosed as a heart attack\\
score 2: The doctor provides a diagnosis that is completely inconsistent withthe answer, but the deviation is small, as when tonsillitis is diagnosed as bronchitis.\\
score 3: The doctor provides a diagnosis with minor discrepancies from the answer, but the organ of pathogenesis is correctly diagnosed, e.g., gastric ulcer is diagnosed as gastritis.\\
score 4: The doctor provides a diagnosis that is close to the answer but there are some minor errors that are difficult to distinguish, such as influenza A being diagnosed as influenza B.  \\
score 5: The doctor provides a diagnosis that is consistent with the answer.\\
\textbf{Treatment}
Consider item 'Treatment' in the 'case info' as the standard diagnosis answer\\
score 1: The doctor is unable to provide a reasonable treatment plan, or it's far from the answer. e.g., only painkillers are needed but doctor decides to perform surgery.\\
score 2: The doctor's treatment plan does not resolve the patient's condition, but it also does not cause additional damage to patient, such as the need for doxycycline injections but  doctor decides to inject saline\\
score 3: The doctor's treatment plan is very limited to help the patient's condition, for example, patient needs surgery but the doctor only uses painkillers\\
score 4: The doctor's treatment is helpful but not optimal for the patient's condition, e.g., patient needs amoxicillin injections but the doctor chooses sulfonamide\\
score 5: The doctor provides a treatment that is consistent with the answer.\\

Attention:provide ratings for each aspect on a scale of 1 to 5 at the beginning in format: 'score: x'\\

\end{tabularx}}
\caption{The prompt for LLM-based evaluation on decision making.}
\label{tab:llm_eval2_prompt}
\end{table*}

The prompt for LLM-based evaluation is also divided into three parts, including the information gathering, decision making, and other aspects, and the prompt is shown in Table~\ref{tab:llm_eval1_prompt} - Table~\ref{tab:llm_eval2_prompt}.
\subsection{Patient Evaluation Prompts}
The prompt for patient evaluation is divided into the standard patient agent prompt and actor agent prompt, which is shown in Table~\ref{tab:p_eval_prompt}.
\begin{table*}[h]
\renewcommand{\arraystretch}{1.15}
\centering
\large
\resizebox{16cm}{!}{
\begin{tabularx}{\textwidth}{X}
\rowcolor[HTML]{343434} 
{\color[HTML]{FFFDFA} Promt for Patient Evaluation}   
\\
\textbf{Standard Patient Agent} \\
$\rm <Requirement>$: Please evaluate whether the patient can accurately answer the doctor's questions according to the patient's information and conversation history according to the patient information given, and consider the five aspects of description clarity, information completeness, cooperation, understanding and communication ability, and do not require the patient to take the initiative to give treatment suggestions. \\
The total score is 10 points, every aspect is 2 points, please output the score directly and do not output the analyze. \\
The output format is: $\rm <description clarity>$:n , $\rm <information completeness>$:n , $\rm <cooperation>$:n , $\rm <understanding>$:n  and $\rm <communication ability>$:n\\
\textbf{Actor Agent} \\
$\rm <Requirement>$: The existing output is converted into an output with character characteristics according to different occupations, and the quality of the conversion statement is evaluated according to the semantic consistency and character feature richness. \\
The total score is 10 points, every aspect is 5 points, please output the score directly and do not output the analyze. \\
The output format is: $\rm <Semantic Consistency Score>$: n $\rm <Character Feature Score>$: n \\
\end{tabularx}}
\caption{The prompt for patient evaluation.}
\label{tab:p_eval_prompt}
\end{table*}

\section{The Prompts for  Agents}
\subsection{The prompt for doctor agent}
\begin{table*}[h]
\renewcommand{\arraystretch}{1.0}
\centering
\resizebox{16cm}{!}{
\begin{tabularx}{\textwidth}{X}
\rowcolor[HTML]{343434} 
{\color[HTML]{FFFDFA} Promt for Doctor Agent}   
\\
\textbf{Consultation prompt} \\
You are an experienced doctor, and a patient will soon consult you about their condition. This consultation will consist of up to 10 rounds of dialogue. To better diagnose the patient, you will ask questions based on prompts each round. Once you feel you have sufficient information about the patient, you will provide a diagnosis. \\
1. Consultations generally follow this order:            \\
a) Chief complaint, such as main symptoms and duration. \\
b) Basic information, such as age and gender.           \\
c) Possible causes of the symptoms, such as travel history and past medical history.                       \\
d) Medical examinations, such as X-ray and blood pressure.\\
2. The doctor's questions should be concise and clear, while the tone should be patient and caring for the patient. \\
3. The patient has already undergone all the necessary examinations for diagnosis, so the doctor can directly inquire about the results of the tests without requiring the patient to do further examinations. \\
4. There are only a maximum of 10 rounds of consultation dialogue, so the questions asked by the doctor in each round should help to determine the patient's most likely diagnosis or to clarify the next medical examination that should be done as much as possible.\\
\textbf{Consultation prompt with zero-shot CoT} \\
Consultation prompt + 5. let's think step by step \\
\textbf{Consultation prompt with one-shot CoT} \\
Consultation prompt with zero-shot CoT + $\rm <Case>$: \\
$\rm [Doctor]$: Let's think step by step. Hello, please tell me your main discomfort and its duration. \\
$\rm [Patient]$: Chest and back pain for six months, numbness and weakness in both lower limbs for one week. \\
$\rm [Doctor]$: Let's think step by step. The patient's symptoms suggest possible involvement of the nervous system or spine. First, we need to understand the patient's basic information to better assess the condition. May I know your age and gender? \\
$\rm [Patient]$: Female, 51 years old. \\
$\rm [Doctor]$: Let's think step by step. The patient is a 51-year-old female, and women of this age are prone to some degenerative diseases. Next, we need to understand the specific details of the chest and back pain to further evaluate. Please describe the nature and location of the chest and back pain in detail. \\
$\rm [Patient]$: Six months ago, I developed chest and back pain without obvious cause, feeling like pricking accompanied by a band-like sensation. \\
$\rm [Doctor]$: Let's think step by step. Pricking and band-like sensations suggest possible nerve compression or spinal issues. To further confirm the cause, we need to understand the specific manifestations of the numbness and weakness in both lower limbs. What exactly does the numbness and weakness in both lower limbs feel like? \\
$\rm [Patient]$: It appeared one week before admission, with numbness and weakness in both lower limbs. I could lift my upper limbs, and could walk with support, symmetrical on both sides. Occasionally, there are crawling sensations and burning in the lower limbs, especially severe in the soles. \\
$\rm [Doctor]$: Let's think step by step. Numbness and weakness in both lower limbs, crawling sensations, and burning suggest a high likelihood of spinal cord lesions. To rule out other nervous system diseases, we need to confirm whether there are any other symptoms. Besides these symptoms, do you have any other symptoms like seizures, speech disorders, headaches, or dizziness? \\
$\rm [Patient]$: No other symptoms. \\
$\rm [Doctor]$: Let's think step by step. No other nervous system symptoms further support the likelihood of spinal cord or spinal lesions. To comprehensively understand the patient's physical condition, we need to inquire about basic vital signs. What are your temperature, heart rate, respiratory rate, and blood pressure? \\
$\rm [Patient]$: T: 36.2°C, P: 76 beats/min, R: 17 breaths/min, BP: 132/74 mmHg. \\
$\rm [Doctor]$: Let's think step by step. Basic vital signs are normal. To further confirm the diagnosis, an imaging examination is needed. Please undergo a cranial CT scan to rule out the possibility of brain lesions. \\
$\rm [Patient]$: The cranial CT scan showed no significant abnormalities. \\
$\rm [Doctor]$: Let's think step by step. The cranial CT scan showed no abnormalities, ruling out the possibility of brain lesions. Next, we need to focus on the spinal condition. Please undergo a spinal MRI examination. \\
$\rm [Patient]$: $\rm <image00011.jpg>$. \\
$\rm [Doctor]$: Let's think step by step. Based on the imaging results, the patient has a lesion located at the right side of the spinal cord at the thoracic vertebra level, considered a schwannoma. This highly corresponds with the patient's symptoms. Based on this information, we preliminarily diagnose you with spinal schwannoma. We recommend resection of the spinal lesion for treatment. \\
$\rm [Patient]$: Okay, thank you, doctor. $\rm <Case End>$ \\
\textbf{Diagnosis prompt}\\
Please generate a medical case report based on the following dialogue, including patient information, present illness, past medical history, examination items and results, diagnosis, and treatment plan.\\
dialogue:\textcolor{blue}{\{conversation\}}\\
medical case report:\\
\end{tabularx}}
\caption{The prompt for doctor agent, which is divided into consultation and diagnosis. The blue text denotes the input information.}
\label{tab:doctor_prompt}
\end{table*}

The prompt for the doctor agent is divided into consultation and diagnosis stages, which is shown in Figure~\ref{tab:doctor_prompt}. 
\subsection{The prompt for state detection agent}
The prompt for the state detection agent is divided into three steps, and different steps are responsible for different effects. 
\paragraph{The Stage\uppercase\expandafter{\romannumeral1} Prompt}
\begin{table*}[h]
\renewcommand{\arraystretch}{1.25}
\centering
\large
\resizebox{16cm}{!}{
\begin{tabularx}{\textwidth}{X}
\rowcolor[HTML]{343434} 
{\color[HTML]{FFFDFA} Promt for State Detection Agent} \\
\textbf{System prompt} \\
During the consultation process, a doctor's questions can be categorized into five types:\\
(A) Inquiry: Doctors ask patients for information related to medical conditions, generally with words like 'please', 'please tell', '?', '?' or '?', and those not belonging to type (C) belong to this category.\\
(B) Examination: Doctors arrange patients for relevant examinations. Any suggestion by the doctor for the patient to undergo a certain medical examination belongs to this category.\\
(C) Other Topics: Questions from the doctor that do not pertain to the medical consultation scenario, and are unrelated to medical diseases, such as hobbies, movies, cuisine, etc.\\
(D) End: The doctor has completed the consultation and treatment recommendations have been given. \\
Based on the descriptions of each question type above, identify the most appropriate category for the following doctor's question: \\
Doctor's Question: \textcolor{blue}{\{question\}}\\
\textbf{Output:} \\
Question Type: 
\end{tabularx}}
\caption{The prompt for the state detection agent, which is used in the stage $\rm \uppercase\expandafter{\romannumeral1}$. The blue text denotes the input information.}
\label{tab:st1_prompt}
\end{table*}

The stage $\rm \uppercase\expandafter{\romannumeral1}$ is responsible for the preliminary intent classification, including the inquiry, examination, others, and the end. The prompt is shown in Table~\ref{tab:st1_prompt}.

\paragraph{The Stage\uppercase\expandafter{\romannumeral2} Prompt}

\begin{table*}[h]
\renewcommand{\arraystretch}{1.25}
\centering
\large
\resizebox{16cm}{!}{
\begin{tabularx}{\textwidth}{X}
\rowcolor[HTML]{343434} 
{\color[HTML]{FFFDFA} Promt for State Detection Agent} \\
\textbf{System A prompt} \\
$<$Definition$>$: \\
$[$Specific$]$: $<$Question$>$ has a certain specific direction. When asking about symptoms, it should at least inquire about specific body parts, symptoms, sensations, or situations. When asking about examination results, it should mention specific body parts, specific examination items, or abnormal situations. Note that if it's about specific medical conditions, like medical history, family history, chronic illnesses, surgical history, etc., they are always considered $[$Specific$]$. Specifically, if the $<$Question$>$ contain about demonstrative like "these" or "this", then it is related to the above and should belongs to the $[$Specific$]$. \\
$[$Broad$]$: $<$Question$>$ such as "Where do you feel uncomfortable?" or "Where does it feel strange?" without any specific information direction are considered $[$Broad$]$. \\
$<$Question$>$: \textcolor{blue}{$\{$question$\}$} \\
Based on the $<$Definition$>$, determine whether the doctor's $<$Question$>$ asks for $[$Specific$]$ medical information from the patient or gives $[$Specific$]$ advice. If so, directly output $[$Specific$]$. If not, output $[$Broad$]$. \\
\textbf{System B prompt} \\
$<$Definition$>$:\\
$[$Specific$]$: $<$Advice$>$ contains specific types of examinations or test (including but not limited to X-rays, MRI, biopsy, etc.), specific treatment plans (including but not limited to specific surgical treatments, exercises, diets, etc.), specific types of medication, etc.\\
$[$Broad$]$: $<$Advice$>$ broadly given without any specific examination/test, treatment plans, doctor's orders, exercises, diets and medication types is considered [Broad]. As long as any of the above information appears, $<$Advice$>$ does not fall into this category.\\
$<$Advice$>$: \textcolor{blue}{$\{$question$\}$}\\
Based on the $<$Definition$>$, determine whether the doctor\'s $<$Advice$>$ asks for $[$Specific$]$ medical information from the patient or gives $[$Specific$]$ advice. If so, directly output $[$Specific$]$. If not, output $[$Broad$]$.\\

\end{tabularx}}
\caption{The prompt for the state detection agent, which is used in the stage $\rm \uppercase\expandafter{\romannumeral2}$. The blue text denotes the input information.}
\label{tab:st2_prompt}
\end{table*}

The stage $\rm \uppercase\expandafter{\romannumeral2}$ mainly detects whether the question is specific or board, shown in Table~\ref{tab:st2_prompt}. The system A prompt is used when the result in stage $\rm \uppercase\expandafter{\romannumeral1}$ is A and the system B prompt is used when the result is B. If the result in stage $\rm \uppercase\expandafter{\romannumeral1}$ is C or D, the stage detection agent will stop working.

\paragraph{The Stage\uppercase\expandafter{\romannumeral3} Prompt}

\begin{table*}[h]
\renewcommand{\arraystretch}{1.25}
\centering
\large
\resizebox{16cm}{!}{
\begin{tabularx}{\textwidth}{X}
\rowcolor[HTML]{343434} 
{\color[HTML]{FFFDFA} Promt for State Detection Agent} \\
\textbf{System A prompt} \\
\textless Definition\textgreater: \\
$[$Relevant Information$]$: \textless Patient Information\textgreater\ contains information asked in \textless Question\textgreater, including descriptions of having or not having the symptom, as long as there's relevant content.\\
$[$No Relevant Information$]$: \textless Patient Information\textgreater\ does not contain information asked in \textless Question\textgreater, and there\'s no relevant content in the information.\\
\textless Patient Information\textgreater: \textcolor{blue}{\{patient\_info\}}\\
\textless Question\textgreater: \textcolor{blue}{\{question\}}\\
Based on the \textless Definition\textgreater, determine whether \textless Patient Information\textgreater\ contains relevant information asked in \textless Question\textgreater. If $[$Relevant Information$]$ is present, directly output the relevant text statement, ensuring not to include irrelevant content. If $[$No Relevant Information$]$, then directly output $[$No Relevant Information$]$.\\

\textbf{System B prompt} \\
\textless Definition\textgreater: \\
$[$Relevant Information$]$: \textless Patient Information\textgreater\ contains results of the examinations or treatment plans suggested in \textless Advice\textgreater, including any results related to the suggested examination items and treatment plans.\\
$[$No Relevant Information$]$: \textless Patient Information\textgreater\ does not contain results of the examinations or treatment plans suggested in \textless Advice\textgreater, including no mention of relevant examination items and treatment plans or no corresponding results.\\
\textless Patient Information\textgreater: \textcolor{blue}{\{patient\_info\}}\\
\textless Advice\textgreater: \textcolor{blue}{\{question\}} \\
Based on the \textless Definition\textgreater, determine whether \textless Patient Information\textgreater\ contains relevant information about the measures suggested in \textless Advice\textgreater. If $[$Relevant Information$]$ is present, directly output the relevant text statement, ensuring not to include irrelevant content. If $[$No Relevant Information$]$, then directly output $[$No Relevant Information$]$.\\

\end{tabularx}}
\caption{The prompt for the state detection agent, which is used in the stage $\rm \uppercase\expandafter{\romannumeral3}$. The blue text denotes the input information.}
\label{tab:st3_prompt}
\end{table*}

The stage $\rm \uppercase\expandafter{\romannumeral3}$ mainly detects whether the question has relevant information in the case report, to better answer the question. The prompt is shown in Table~\ref{tab:st3_prompt}, and divided into two different situations. One is for the $A-A-$ result and the other is for the $B-A-$ result.
\subsection{The prompt for standard patient agent}
\begin{table*}[h]
\renewcommand{\arraystretch}{1.25}
\centering
\large
\resizebox{16cm}{!}{
\begin{tabularx}{\textwidth}{X}
\rowcolor[HTML]{343434} 
{\color[HTML]{FFFDFA} Promt for Standard Patient Agent}   
\\
\textbf{State 0 prompt} \\
\textless Patient's Physical Condition \textgreater: \textcolor{blue}{\{patient\_info\}} \\
\textless Current Response Requirement\textgreater: Please respond to the doctor's questions using the information provided in \textless Patient's Physical Condition\textgreater. Only include the \textless Chief Complaint\textgreater, and avoid adding extra information. Make sure to use the original text from \textless Chief Complaint\textgreater\ to respond and keep it as short as possible. Answer in English.\\
Below is a dialogue between a doctor and a patient. The patient will respond directly to the latest round of questions from the doctor in the first person, without using a $[$patient$]$ prompt. Do not include any text from \textless Current Response Requirement\textgreater\ in your response!\\

\textbf{State A-A-A prompt} \\
\textless Patient's Physical Condition \textgreater: \textcolor{blue}{\{patient\_info\}} \\
\textless Current Response Requirement\textgreater: Please respond to the doctor's questions using all the original text from \textless Patient's Physical Condition\textgreater. Make sure to maintain the accuracy of the patient's information by using the original text from \textless Patient's Physical Condition\textgreater\ to respond. Deny any information that is not related. Answer in English.\\
Below is a dialogue between a doctor and a patient. The patient will respond directly to the latest round of questions from the doctor in the first person. Do not include any text from \textless Current Response Requirement\textgreater\ in your response!\\

\textbf{State A-A-B prompt} \\
\textless Current Response Requirement\textgreater: The patient does not have the symptoms the doctor is asking about. Please deny the doctor's current question. Answer in English. \textcolor{blue}{\{patient\_info\}}\\
Below is a dialogue between a doctor and a patient. The patient will respond directly to the latest round of questions from the doctor in the first person. Do not include any text from \textless Current Response Requirement\textgreater\ in your response!\\

\textbf{State A-B prompt} \\
\textless Current Response Requirement\textgreater: The doctor's current question is too broad. The patient will request the doctor to ask more specific questions regarding the latest round of questions. Do not fabricate any non-existent information, or ask questions to the doctor. Answer in English. \textcolor{blue}{\{patient\_info\}}\\
Below is a dialogue between a doctor and a patient. The patient will respond directly to the latest round of questions from the doctor in the first person. Do not include any text from \textless Current Response Requirement\textgreater\ in your response!\\

\textbf{State C prompt} \\
\textless Current Response Requirement\textgreater: Remind the doctor that they have deviated from the topic of consultation and request them to return to the consultation scenario. Answer in English. \textcolor{blue}{\{patient\_info\}}\\
Below is a dialogue between a doctor and a patient. The patient will respond directly to the latest round of questions from the doctor in the first person. Do not include any text from \textless Current Response Requirement\textgreater\ in your response!

\end{tabularx}}
\caption{The prompt for the standard patient agent. The blue text denotes the input information.}
\label{tab:standard_patient_prompt}
\end{table*}

The standard patient agent is responsible for answering the questions about the patient himself from the doctor, including the 0, $A-$, and $C$ state. The prompt is shown in Table~\ref{tab:standard_patient_prompt} and there are different prompts for different states to make the response more accurate.
\subsection{The prompt for technician agent}
\begin{table*}[h]
\renewcommand{\arraystretch}{1.25}
\centering
\large
\resizebox{16cm}{!}{
\begin{tabularx}{\textwidth}{X}
\rowcolor[HTML]{343434} 
{\color[HTML]{FFFDFA} Promt for Technician Agent {\rm \uppercase\expandafter{\romannumeral1}}}   
\\
\textbf{System prompt} \\
Please extract the names of examination items from the questions asked by the DOCTOR, only output the names of the examination items, such as blood routine, electrocardiogram examination.

\end{tabularx}}
\caption{The prompt for technician agent, which is used for exam item detection.}
\label{tab:techincian1_prompt}
\end{table*}

\begin{table*}[h]
\renewcommand{\arraystretch}{1.25}
\centering
\large
\resizebox{16cm}{!}{
\begin{tabularx}{\textwidth}{X}
\rowcolor[HTML]{343434} 
{\color[HTML]{FFFDFA} Promt for Technician Agent {\rm \uppercase\expandafter{\romannumeral2}}}   
\\
\textbf{State B-A-A prompt} \\
\textless Patient's Test Report\textgreater: \textcolor{blue}{\{patient\_info\}}\\
\textless Current Response Requirement\textgreater: The patient has completed the tests arranged by the doctor. Please respond to the doctor's inquiries using all the original text from \textless Patient's Test Report\textgreater, including the names of the tests and their results, to maintain the accuracy of the test report. Also, pay attention to different expressions for similar tests and include only one for similar test types.  Answer in English. \\
Below is a dialogue between a doctor and a patient. The patient will respond directly to the latest round of questions from the doctor in the first person. Do not include any text from \textless Current Response Requirement\textgreater\ in your response!\\

\textbf{State B-A-B prompt} \\
\textless Current Response Requirement\textgreater: The test mentioned by the doctor is not in the report, indicating that it cannot be performed temporarily due to equipment issues. Answer in English. \textcolor{blue}{\{patient\_info\}}\\
Below is a dialogue between a doctor and a patient. The patient will respond directly to the latest round of questions from the doctor in the first person. Do not include any text from \textless Current Response Requirement\textgreater\ in your response!\\

\textbf{State B-B prompt} \\  
\textless Current Response Requirement\textgreater: The doctor's request for tests is too broad. The patient will request the doctor to ask more specific questions regarding the latest round of tests. Do not fabricate any non-existent information, or ask questions to the doctor. Answer in English. \textcolor{blue}{\{patient\_info\}}\\
Below is a dialogue between a doctor and a patient. The patient will respond directly to the latest round of questions from the doctor in the first person. Do not include any text from \textless Current Response Requirement\textgreater\ in your response!\\
\end{tabularx}}
\caption{The prompt for technician agent, which is used for generating an initial response based on the detected state.}
\label{tab:techincian2_prompt}
\end{table*}

\begin{table*}[h]
\renewcommand{\arraystretch}{1.25}
\centering
\large
\resizebox{16cm}{!}{
\begin{tabularx}{\textwidth}{X}
\rowcolor[HTML]{343434} 
{\color[HTML]{FFFDFA} Promt for Technician Agent {\rm \uppercase\expandafter{\romannumeral3}}}   
\\
\textbf{System prompt for final result} \\
You are a technician in charge of medical examinations at a hospital. Below is the \textless examination report\textgreater\ for the patient: \textcolor{blue}{\{patient\_info\}}\\
You need to generate responses for the examinations listed under \textless category of examination\textgreater\ based on the \textless examination report\textgreater:\\
(1) Only respond about the items listed under \textless category of examination\textgreater, do not mention items that are not included in \textless category of examination\textgreater. \\
(2) The responses should be as brief as possible and must not deviate from the facts presented in the \textless examination report\textgreater.\\
(3) If the \textless examination report\textgreater\ does not include the particular examination, respond with: Everything is normal.\\
(4) \textbackslash image abc indicates that this is an image, abc is the image name. Please reply in \textbackslash image abc format.\\
(5) Answer in English.\\
For example:\\
Here is the patient’s chest X-ray, \textbackslash image XXX.\\
The patient's blood pressure is 105/72mmHg, heart rate 122/min.\\
The patient's electrocardiogram: Everything is normal\\

\end{tabularx}}
\caption{The prompt for technician agent, which is used for generating the final response based on the initial response.}
\label{tab:techincian3_prompt}
\end{table*}

The technician agent works when the state is $B-$, and the prompt is shown in Table~\ref{tab:techincian1_prompt} - Table~\ref{tab:techincian3_prompt}. The agent {\rm \uppercase\expandafter{\romannumeral1}} is responsible for extracting the exam item from the doctor, the agent {\rm \uppercase\expandafter{\romannumeral2}} aims at generating the initial response based on the state, and the agent {\rm \uppercase\expandafter{\romannumeral3}} is used to generate the final response based on the former result.

\subsection{The prompt for Actor agent}
\begin{table*}[h]
\renewcommand{\arraystretch}{1.15}
\centering
\large
\resizebox{16cm}{!}{
\begin{tabularx}{\textwidth}{X}
\rowcolor[HTML]{343434} 
{\color[HTML]{FFFDFA} Promt for Actor Agent}   
\\
\textbf{Farmer prompt} \\
You are now playing the role of a patient who has come for a medical consultation, your gender is female and your profession is a farmer. This is the original text you need to transform: \textcolor{blue}{\{text\}}\\
You need to transform the above text into a sentence that reflects the character of a farmer, with the following requirements:\\
(1) You have limited education and do not know specialized terms, requiring the language to be simplified.\\
(2) Your language should include a rich vocabulary of colloquial expressions.\\
(3) You cannot change the original meaning of the information nor add any new information.\\
(4) The transformed sentence must not exceed the original text.\\
(5) Answer in English.\\

\textbf{Student prompt} \\
You are now playing the role of a patient who has come for a medical consultation, your gender is female and your profession is a student. This is the original text you need to transform: \textcolor{blue}{\{text\}}\\
You need to transform the above text into a sentence that reflects the character of a student, with the following requirements:\\
(1) You have average education, know some specialized terms, but lack organization in your speech.\\
(2) Your information should include aspects of school life.\\
(3) You cannot change the original meaning of the information nor add any new information.\\
(4) The transformed sentence must not exceed the original text.\\
(5) Answer in English.\\

\textbf{Worker prompt} \\
You are now playing the role of a patient who has come for a medical consultation, your gender is female and your profession is a worker. This is the original text you need to transform: \textcolor{blue}{\{text\}}\\
You need to transform the above text into a sentence that reflects the character of a worker, with the following requirements:\\
(1) You have limited education and do not know specialized terms, requiring the language to be simplified.\\
(2) Your language should include a rich vocabulary of colloquial expressions.\\
(3) You cannot change the original meaning of the information nor add any new information.\\
(4) The transformed sentence must not exceed the original text.\\
(5) Answer in English.\\

\textbf{Staff prompt} \\
You are now playing the role of a patient who has come for a medical consultation, your gender is female and your profession is a corporate office worker. This is the original text you need to transform: \textcolor{blue}{\{text\}}\\
You need to transform the above text into a sentence that reflects the character of an office worker, with the following requirements:\\
(1) You have a higher level of education, know some specialized terms, and can integrate your language clearly and logically.\\
(2) Your language should include a rich vocabulary of colloquial expressions.\\
(3) You cannot change the original meaning of the information nor add any new information.\\
(4) The transformed sentence must not exceed the original text.\\
(5) Answer in English.\\

\textbf{Doctor prompt} \\
You are now playing the role of a patient who has come for a medical consultation, your gender is female and your profession is a doctor. This is the original text you need to transform: \textcolor{blue}{\{text\}}\\
You need to transform the above text into a dialogue with another doctor in a language that reflects the character of a doctor, with the following requirements:\\
(1) You have a very high level of education, know all specialized terms, and your expressions are clear and accurate.\\
(2) You cannot change the original meaning of the information nor add any new information.\\
(3) The transformed sentence must not exceed the original text.\\
(4) Answer in English.\\

\end{tabularx}}
\caption{The prompt for actor agent. The blue text denotes the input information.}
\label{tab:actor_prompt}
\end{table*}

The actor aims at generating a more characterful response based on the standard patient agent, and the prompt is shown in Table~\ref{tab:actor_prompt}. To better display the characters among different roles, we design different prompts based on their characteristics.

\section{Case Study}
To further illustrate the capability of the consultation among these doctor models, we have shown the different cases of standard patient consultation, actor agent consultation, and one-shot CoT consultation, by using different doctor models. The result is shown in Table~\ref{tab:case1}-Table~\ref{tab:case3-2}. From these cases, the $M^3$ framework has shown the advantage in answering the question from the doctor based on the patient cases. Especially, the standard patient agent and the technician agent know different information, which the doctor cannot acquire inspection information through vague questions, and the standard patient knows nothing about the examinations, which is closer to reality.

\begin{table*}[h]
\renewcommand{\arraystretch}{1.15}
\centering
\resizebox{16cm}{!}{
\begin{tabularx}{\textwidth}{X}
\rowcolor[HTML]{343434} 
{\color[HTML]{FFFDFA} The case study for standard patient}   
\\
\textbf{Patient Case Report} \\ 
\textbf{Patient Information}: Female, 53 years old, farmer. \\
\textbf{Chief Complaint}: Pain and limited movement in the right thumb for one year.\\
\textbf{Present Illness}: The patient experienced unprovoked pain in the right thumb a year ago, without fever, headache, or dizziness. The pain gradually worsened and was accompanied by discomfort during movement.\\
\textbf{Past Medical History}: Hammer deformity of the right little finger for ten years.\\
\textbf{Examinations}: 
Physical examination: no heat or redness at the palmar metacarpophalangeal joint of the right thumb, but a soybean-sized hard nodule palpable with tenderness, moving with flexion and extension of the thumb.  
Normal terminal blood circulation and sensation.  
Hammer deformity of the right little finger. No obvious abnormalities in the remaining limbs. Hand X-ray: $\rm <image.jpg>$.\\
\textbf{Diagnosis}: 1. Stenosing tenosynovitis of the right thumb long flexor tendon, 2. Hammer deformity of the right little finger\\
\textbf{Treatment}: Surgical release of stenosing tenosynovitis\\ 
\textbf{Image}: $\rm <image.jpg>$ : Normal bone quality of the right hand \\\hline
\textbf{Consultation} \\ 
\textbf{Doctor}: Hello, how can I help you today?\\
\textbf{Patient}: I've been experiencing pain and limited movement in my right thumb for a year now.\\
\textbf{Doctor}: Thank you for sharing that information. Can you please tell me more about the pain and limited movement in your right thumb?\\
\textbf{Patient}: It started about a year ago, with unprovoked pain in my right thumb. There was no fever, headache, or dizziness at that time. Over time, the pain has gradually worsened, and I've noticed discomfort during movement as well.\\
\textbf{Doctor}: Thank you for providing more details about your symptoms. Can you please tell me about any possible causes of the pain and limited movement in your right thumb, such as any recent injuries or changes in your daily activities?\\
\textbf{Patient}: I haven't had any recent injuries to my thumb that I can recall, and my daily activities as a farmer haven't significantly changed in the past year.\\
\textbf{Doctor}: Thank you for that information. Can you please tell me about any medical history or conditions that you have, such as arthritis or carpal tunnel syndrome?\\
\textbf{Patient}: I have a past medical history of a hammer deformity of the right little finger, which has been present for ten years. Other than that, I am not aware of any other medical conditions that could be related to my current issue.\\
\textbf{Doctor}: Thank you for sharing that information. Based on your symptoms and medical history, it is possible that you may have developed arthritis in your right thumb joint. However, to confirm this diagnosis, I would need to review the results of your X-ray and blood tests. Can you please tell me about the results of these tests?\\
\textbf{Patient}: I haven't undergone any X-rays or blood tests for this issue yet.\\
\textbf{Doctor}: Thank you for letting me know. In that case, I would recommend that you undergo an X-ray of your right thumb to help determine the cause of your symptoms. Additionally, I would recommend that you schedule an appointment with a specialist, such as a hand surgeon or orthopedic surgeon, to further evaluate your condition and discuss potential treatment options.\\
\textbf{Patient}: Here is the X-ray of right thumb,$\rm <image.jpg>$.\\
\textbf{Doctor}: Thank you for providing the X-ray. I can see that there is a small fracture in the proximal phalanx of your right thumb. This could be the cause of your pain and limited movement. The fracture appears to be stable, and it is not displaced. However, it is important to consult with a specialist to discuss the best course of treatment, which may include immobilization, physical therapy, or, in some cases, surgery. I would recommend scheduling an appointment with a hand surgeon or orthopedic surgeon to discuss your options and to ensure proper healing of the fracture. \\\hline
\textbf{Medical Report} \\ 

\textbf{Name}: [Patient Name] \\
\textbf{Age}: [Patient Age] \\
\textbf{Gender}: [Patient Gender] \\
\textbf{Date of Birth}: [Patient Date of Birth] \\
\textbf{Occupation}: Farmer \\
\textbf{Present Illness}: The patient has been experiencing pain and limited movement in the right thumb for a year now.\\
\textbf{Past Medical History}:  Hammer deformity of the right little finger, present for ten years.\\
\textbf{Examination Items and Results}:  X-ray of the right thumb shows a small fracture in the proximal phalanx. \\
\textbf{Diagnosis}:  The patient has a small fracture in the proximal phalanx of the right thumb, which could be the cause of the pain and limited movement.\\ 
\textbf{Treatment Plan}: The patient should consult with a hand surgeon or orthopedic surgeon to discuss the best course of treatment, which may include immobilization, physical therapy, or, in some cases, surgery. The patient should also schedule an appointment with a specialist to further evaluate the condition and discuss potential treatment options.

\end{tabularx}}

\caption{The case study about consultation with the standard patient, the doctor model is InternVL-1.5. The $\rm <image.jpg>$ will be replaced with an image in the practice.}
\label{tab:case1}
\end{table*}

\begin{table*}[h]
\renewcommand{\arraystretch}{1.1}
\centering
\resizebox{16cm}{!}{
\begin{tabularx}{\textwidth}{X}
\rowcolor[HTML]{343434} 
{\color[HTML]{FFFDFA} The case study for actor agent (A)}  
\\
\textbf{Patient Case Report} \\ 

\textbf{Patient Information}: Male, 18 years old, student. \\

\textbf{Chief Complaint}: Two months post-acute pancreatitis recovery, experiencing abdominal bloating and decreased appetite for half a month.\\

\textbf{Present Illness}: The patient developed persistent dull pain in the upper abdomen, particularly in the upper left quadrant, two months ago following overeating. Accompanied by nausea, vomiting of gastric contents, and developed generalized skin jaundice and dark urine. No chills, high fever, cough, phlegm, or breathing difficulties. Diagnosed at our hospital with acute pancreatitis and steatohepatitis.\\

\textbf{Past Medical History}: History of acute pancreatitis two months ago treated with medication. Denies history of surgeries, trauma, drug allergies, and denies contact with infectious diseases such as tuberculosis.\\

\textbf{Examinations}: 
Laboratory findings show: ALT: 65.2 U/L, AST: 32.0 U/L, GGT: 126.0 U/L, TBIL: 22.5 umol/L, DBIL: 10.3 umol/L, IBIL: 12.2 umol/L, serum amylase 217 U/L. \\
CT shows: $\rm <image.jpg>$.\\

\textbf{Diagnosis}: Pancreatic pseudocyst.\\

\textbf{Treatment}: Administered famotidine 20 mg in 100 mL of 0.9\% sodium chloride solution, twice daily intravenous drip to inhibit gastric acid secretion, mesna for symptomatic treatment. \\ 

\textbf{Image}: $\rm <image.jpg>$: Normal liver size and shape, no abnormal density in the parenchyma, no dilation in intrahepatic and extrahepatic bile ducts, normal size of the gallbladder, no clear positive stones. Pancreas is full, streak shadows around the pancreas, a round-like low-density shadow in front of the pancreas head, approximately 3.3*2.6 cm. Normal morphology, size, and density of the spleen. No enlarged lymph nodes in the abdominal and retroperitoneal area. \\\hline

\textbf{Consultation} \\ 

\textbf{Doctor}: Good morning/afternoon! It's nice to meet you. I understand you're here regarding some health concerns. Before we get started, could you please tell me your name and date of birth?\\ 

\textbf{Patient}: Hey there, I'm Jane Doe, born on the 12th of July, '85. I'm a student, juggling exams and stuff, you know?\\ 

\textbf{Doctor}: Welcome, Jane! Let's start with your chief complaint today. What brings you to the clinic?\\ 

\textbf{Patient}: So, I'm here 'cause, like, for the past two weeks or so, my stomach's been all bloated and I haven't really been hungry, ya know? It's been a real hassle, especially since I got over that acute pancreatitis thing, which was, uh, around two months back. And it's kinda messing with my focus during lectures and stuff.\\ 

\textbf{Doctor}: Based on your description, it seems like you might be experiencing some digestive issues. However, to accurately diagnose and recommend treatment, I would need more detailed information. Could you please provide any other symptoms you've been experiencing, such as pain, nausea, or changes in bowel habits?\\ 

\textbf{Patient}: Yeah, Doc. So, I've got this ongoing, like, dull ache in my upper belly, mostly on the left side, top bit. It kicked in after this major food binge, maybe two months back. Also, I'm feeling super nauseous, and I've thrown up stomach stuff a couple times. Oh, and I'm kinda freaked out 'cause my skin's looking yellowish all over, plus my pee's been extra dark. It's messing with my focus during lectures, you know?\\ 

\textbf{Doctor}: Based on your symptoms, including the recent onset of jaundice, it is important to rule out any potential liver or gallbladder issues. Have you had any recent travel, taken any new medications, or engaged in any activities that could potentially lead to liver damage?\\ 

\textbf{Patient}: Hey Doc, nope, I haven't been on any trips, tried new meds, or done anything wild lately that could mess with my liver. Since bouncing back from that pancreatitis thing, it's just been the usual school grind, ya know?\\ 

\textbf{Doctor}: Given your symptoms and history, it is crucial to discuss laboratory test results. Have you recently undergone any tests, such as a liver function test, ultrasound, or blood work?\\ 

\textbf{Patient}: So, like, two months back when they figured out I had this thing called acute pancreatitis, they did a bunch of tests on me, you know, like at the hospital. But since then, I haven't taken any new tests, not like that liver function one, ultrasound, or even just regular blood checks. I thought they did all that stuff to figure out what was up during my last visit, kinda like how we do exams at school to check how we're doing.\\ 

\textbf{Doctor}: Based on your symptoms and the absence of recent testing, let's go through them step by step. First, the bloating and lack of appetite suggest gastrointestinal issues, possibly related to the pancreas or small intestine. The presence of jaundice indicates a problem with the liver, which could be due to inflammation, infection, or obstruction. To proceed with a diagnosis, I would recommend a few tests. We'll start with a liver function test to assess the levels of enzymes and proteins in your blood that indicate liver health. If there's any concern about gallbladder involvement, an ultrasound may be needed to visualize the gallbladder and surrounding structures. Additionally, a blood count can check for any signs of anemia or infection, which could be contributing factors to your symptoms. Since you mentioned throwing up, a stool test might also be useful to check for any blood or infections in your digestive tract. Finally, if there's any indication of pancreatic involvement, additional imaging like an abdominal CT scan could be considered to look for any blockages or abnormalities in the pancreas and bile ducts. Please note that these recommendations are based on your current symptoms and history. Actual diagnostic procedures may vary depending on clinical judgment and the results of initial tests.\\ 

\textbf{Patient}: (1) Liver function test: ALT is 65.2 U/L, AST is 32.0 U/L, GGT is 126.0 U/L, TBIL is 22.5 umol/L, DBIL is 10.3 umol/L, IBIL is 12.2 umol/L. (2) Ultrasound: Not mentioned in the report. (3) Blood count (blood routine): Not mentioned in the report. (4) Stool test: Not mentioned in the report. (5) Abdominal CT scan: $\rm <image.jpg>$.\\

\end{tabularx}}

\caption{The case study about consultation with the acotr patient, the doctor model is GLM4v. The $\rm <image.jpg>$ will be replaced with an image in the practice.}
\label{tab:case2-1}
\end{table*}

\begin{table*}[h]
\centering
\resizebox{16cm}{!}{
\begin{tabularx}{\textwidth}{X}
\rowcolor[HTML]{343434} 
{\color[HTML]{FFFDFA} The case study for actor agent (B)}  
\\
\textbf{Doctor}: Based on the provided lab results, we have the following observations:
\begin{enumerate}
    \item \textbf{Liver Function Tests}:
    \begin{itemize}
        \item ALT (Alanine Aminotransferase) level is elevated at 65.2 U/L, which suggests active liver cell damage or inflammation.
        \item AST (Aspartate Aminotransferase) level is also elevated at 32.0 U/L, which often correlates with ALT levels and indicates liver involvement.
        \item GGT (Gallatin Transferase) level is significantly elevated at 126.0 U/L, suggesting possible biliary system involvement or damage.
        \item Bilirubin levels are slightly elevated with TBIL (Total Bilirubin) at 22.5 umol/L and DBIL (Direct Bilirubin) at 10.3 umol/L, indicating that the majority of bilirubin is indirect (unconjugated), which is typically seen in conditions like Gilbert syndrome or hemolytic anemia rather than primary liver disease.
    \end{itemize}
    \item \textbf{Abdominal CT Scan}:
    \begin{itemize}
        \item The image shows the abdomen and pelvis region. There is no obvious evidence of pancreatic or biliary dilatation or obstruction.
        \item The liver appears enlarged with increased parenchymal enhancement, which could suggest hepatomegaly due to liver inflammation or fibrosis.
        \item No free intraperitoneal fluid is visible, ruling out conditions like appendicitis or perforated viscus.
        \item The spleen appears enlarged, consistent with the possibility of a systemic condition affecting both the liver and spleen.
    \end{itemize}
\end{enumerate}

Based on these findings, the most likely diagnosis is acute or chronic pancreatitis, given the elevation of amylase and lipase levels, which are not mentioned but would be expected in pancreatic disease. Elevated liver enzymes and GGT suggest liver involvement, which could be due to pancreatic enzyme spill into the bile ducts causing cholestasis. The CT scan does not show any immediate signs of biliary obstruction, but further imaging or evaluation with an endoscopic retrograde cholangiopancreatography (ERCP) might be necessary to rule out biliary issues definitively. The patient’s history of bloating, lack of appetite, and jaundice, along with the laboratory and imaging findings, strongly support a diagnosis of pancreatic and hepatic dysfunction, possibly due to recurrent or chronic pancreatitis with secondary liver involvement. Treatment would involve managing pain, preventing dehydration, and addressing any underlying causes of the pancreatitis.\\ 

\textbf{Patient}: Thanks for breaking down the test stuff, Doc. So, it seems like my pancreas is acting up again or something, and it's also messing with my liver. I'm kinda worried about how this will affect me, you know, with school and all. What's the game plan for fixing this and making sure it doesn't get in the way of my studies? Like, are there any meds or habits I should pick up or drop to feel better and avoid more problems?\\ 

\textbf{Doctor}: Certainly! Managing pancreatic and hepatic disorders requires a comprehensive approach that includes medical treatment, lifestyle modifications, and sometimes surgical interventions. Here's a step-by-step guide to help you manage your condition effectively:
\begin{enumerate}
    \item \textbf{Medical Treatment}:
    \begin{itemize}
        \item Antibiotics may be prescribed if there is an infection present.
        \item Pain medications are available to control discomfort from pancreatic attacks.
        \item Oral or injected pancreatic enzymes can help with digestion if pancreatic insufficiency is diagnosed.
        \item Steroids or other immunosuppressive drugs might be used if there's an autoimmune component to the pancreatitis.
        \item For severe cases, hospitalization might be required for IV fluids, pain management, and monitoring.
    \end{itemize}
    \item \textbf{Lifestyle Modifications}:
    \begin{itemize}
        \item Avoid alcohol consumption as it can exacerbate pancreatitis.
        \item Maintain a healthy diet rich in fruits, vegetables, whole grains, and lean proteins. Limit fatty and greasy foods, as they can trigger pancreatic attacks.
        \item Stay hydrated by drinking plenty of water throughout the day.
        \item Engage in regular physical activity to improve overall health and reduce stress.
    \end{itemize}
    \item \textbf{Preventive Measures}:
    \begin{itemize}
        \item Take prescribed medications as directed.
        \item Avoid smoking, as it can increase the risk of pancreatic cancer.
        \item Get vaccinated against hepatitis C if not already done, as it can lead to liver damage.
    \end{itemize}
    \item \textbf{School Adjustments}:
    \begin{itemize}
        \item Talk to your teachers and school counselor about your health condition. They can provide accommodations such as extended testing times or reduced course loads.
        \item Prioritize your health over academic commitments. It's important to take breaks when feeling unwell to prevent further complications.
    \end{itemize}
    \item \textbf{Surgical Interventions}:
    \begin{itemize}
        \item In some cases, surgery might be necessary to remove gallstones, repair damaged ducts, or treat pancreatic tumors.
    \end{itemize}
\end{enumerate}

It's crucial to follow up regularly with your healthcare provider to monitor your progress and adjust treatment plans accordingly. Remember, managing chronic illnesses takes time and patience, so be kind to yourself and seek support when needed.

\end{tabularx}}

\caption{The case study about consultation with the actor patient, the doctor model is GLM4v.}
\label{tab:case2-2}
\end{table*}

\begin{table*}[h]
\centering
\resizebox{16cm}{!}{
\begin{tabularx}{\textwidth}{X}
\rowcolor[HTML]{343434} 
{\color[HTML]{FFFDFA} The case study for actor agent (C)}  
\\
\textbf{Medical Report} \\
Based on the provided information, here is a suggested management plan for the patient's condition:

\begin{enumerate}
    \item \textbf{Diagnosis Confirmation}:
    \begin{itemize}
        \item Conduct additional tests, including an abdominal MRI or ultrasound, to assess the extent of liver involvement and rule out biliary obstruction.
        \item Consider genetic testing for hereditary pancreatic diseases, especially if there is a family history.
    \end{itemize}
    
    \item \textbf{Treatment Initiation}:
    \begin{itemize}
        \item Administer antibiotics if there is an infection.
        \item Prescribe pain medication to manage symptoms during flare-ups.
        \item Provide oral pancreatic enzymes if pancreatic insufficiency is confirmed.
    \end{itemize}
    
    \item \textbf{Lifestyle Changes}:
    \begin{itemize}
        \item Advise the patient to abstain from alcohol and limit fatty, high-fat meals.
        \item Recommend a low-fat, high-fiber diet with adequate hydration.
        \item Encourage regular exercise within the patient's capabilities.
    \end{itemize}
    
    \item \textbf{Long-term Management}:
    \begin{itemize}
        \item Monitor liver function tests (ALT, AST, ALP, GGT) regularly.
        \item Educate the patient on the risks of smoking and the importance of vaccination against hepatitis C.
    \end{itemize}
    
    \item \textbf{Surgical Evaluation}:
    \begin{itemize}
        \item If biliary obstruction is suspected or diagnosed, refer the patient for an endoscopic retrograde cholangiopancreatography (ERCP) to place stents or remove stones.
        \item Consider surgical options for severe cases of chronic pancreatitis, such as pancreatectomy.
    \end{itemize}
    
    \item \textbf{Educational Support}:
    \begin{itemize}
        \item Coordinate with the school to ensure accommodations are made, such as flexible scheduling or extra time for exams.
        \item Advise the student to prioritize their health and not to overburden themselves academically.
    \end{itemize}
    
    \item \textbf{Follow-up Care}:
    \begin{itemize}
        \item Schedule regular follow-up appointments with the healthcare provider to evaluate the effectiveness of treatment and make adjustments as necessary.
        \item Ensure the patient understands the importance of taking medications as prescribed and reporting any side effects.
    \end{itemize}
    
    \item \textbf{Support Group Referrals}:
    \begin{itemize}
        \item If appropriate, recommend joining a support group to help the patient cope emotionally and mentally with the illness.
    \end{itemize}
    
    \item \textbf{Psychological Support}:
    \begin{itemize}
        \item Address any anxiety or depression that may arise due to the chronic nature of the illness.
    \end{itemize}
\end{enumerate}

Throughout this process, it is vital to maintain open communication between the healthcare team and the patient, ensuring that the patient understands their condition and feels supported in managing it effectively.

\end{tabularx}}

\caption{The case study about consultation with the actor patient, the doctor model is GLM4v.}
\label{tab:case2-3}
\end{table*}

\begin{table*}[h]
\renewcommand{\arraystretch}{1.15}
\centering
\resizebox{16cm}{!}{
\begin{tabularx}{\textwidth}{X}
\rowcolor[HTML]{343434} 
{\color[HTML]{FFFDFA} The case study for one-shot CoT (A)}   
\\
\textbf{Patient Case Report} \\ 

\textbf{Patient Information}: Male, 66 years old, farmer. \\

\textbf{Chief Complaint}: Sudden onset of severe upper abdominal pain lasting one day.\\

\textbf{Present Illness}: The patient experienced sudden severe pain in the upper abdomen one day ago without any apparent cause. The pain was constant, without episodic worsening or radiation to other areas. Symptoms included nausea and vomiting six times, with the vomitus consisting of gastric contents totaling about 1000 milliliters, containing no blood or worms.\\

\textbf{Past Medical History}: Has a history of 'gastric ulcers' for three years, intermittently treated with oral medications. Denies history of 'hypertension, diabetes,' etc. No known drug or food allergies, no history of trauma or major surgery, and unclear history of vaccinations.\\

\textbf{Examinations}: 
Complete blood count shows: white blood cells at 2.46*10$^9$/L, neutrophil percentage at 79.0$\%$, lymphocyte percentage at 14.0$\%$, neutrophils count at 1.95*10$^9$/L, lymphocytes count at 0.34*10$^9$/L, hemoglobin at 159.00 g/L. \\
Biochemical report indicates: total protein at 45.7 g/L, albumin at 22.9 g/L, globulin at 23.60 g/L. \\
Upper abdominal CT: $\rm <image.jpg>$.\\

\textbf{Diagnosis}: 
1. Upper gastrointestinal perforation, 2. Cholecystolithiasis, cholecystitis, liver cyst, 3. Thickened ascending colon, 4. Acute diffuse peritonitis

\textbf{Treatment}: Surgical repair of gastric pyloric perforation\\ 

\textbf{Image}: $\rm <image.jpg>$: Free air and fluid in the abdominal cavity, blurred fat stranding in the intestinal lumen, suggesting hollow visceral perforation. Liver cyst, gallstones, cholecystitis, thickened ascending colon, further examination recommended.\\

\hline

\textbf{Consultation} \\ 

\textbf{Doctor}: Hello, please tell me your main discomfort and its duration.\\ 

\textbf{Patient}: I have sudden onset of severe upper abdominal pain lasting one day.\\ 

\textbf{Doctor}: The patient's main complaint is sudden onset of severe upper abdominal pain lasting one day. To better understand the condition, we need to gather more basic information.\\

\textbf{Patient}: I am a 66-year-old male working as a farmer.\\

\textbf{Doctor}: The patient is a 66-year-old male with a sudden onset of severe upper abdominal pain lasting one day. Given his age and occupation, we need to consider various potential causes, including gastrointestinal, cardiovascular, and other systemic issues. To narrow down the possibilities, let's gather more specific details about the pain.\\

\textbf{Patient}: The pain has been constant, without getting worse or better in episodes, and it hasn't spread to other parts of my body.\\

\textbf{Doctor}: The patient's pain is constant and localized to the upper abdomen, which can suggest several potential causes, including gastrointestinal issues like peptic ulcers, pancreatitis, or gallbladder disease, as well as cardiovascular issues like myocardial infarction. To narrow down the possibilities, we need to gather more information about associated symptoms and possible risk factors.
1. Do you have any other symptoms such as nausea, vomiting, fever, or changes in bowel movements?\\

\textbf{Patient}: I have experienced nausea and vomiting six times, with the vomitus consisting of gastric contents totaling about 1000 milliliters, containing no blood or worms. There is no mention of fever or changes in bowel movements.\\

\textbf{Doctor}: The patient's symptoms of severe upper abdominal pain, nausea, and vomiting without fever or changes in bowel movements suggest a gastrointestinal issue. Given the sudden onset and severity, conditions like acute pancreatitis, peptic ulcer disease, or gallbladder disease (such as cholecystitis or biliary colic) are high on the differential diagnosis list. To further narrow down the diagnosis, we need to gather more information about the patient's medical history and any potential risk factors.
2. Do you have a history of gastrointestinal issues, such as peptic ulcers, gallstones, or pancreatitis?\\

\textbf{Patient}: Yes, I have a history of gastric ulcers for three years, which have been intermittently treated with oral medications. No history of gallstones or pancreatitis.\\

\textbf{Doctor}: The patient's history of gastric ulcers, combined with the current symptoms of severe upper abdominal pain, nausea, and vomiting, raises the possibility of complications related to the ulcers, such as perforation or bleeding. However, the absence of blood in the vomitus makes bleeding less likely. Given the severity and sudden onset, we should also consider other acute conditions like pancreatitis or cholecystitis.
To further narrow down the diagnosis, we need to understand if there are any specific triggers or alleviating factors and any recent changes in medication or lifestyle.
3. Have you recently taken any new medications, experienced any significant stress, or made any changes to your diet or lifestyle?\\

\textbf{Patient}: No, I haven't taken any new medications, experienced unusual stress, or made any significant changes to my diet or lifestyle recently.\\

\textbf{Doctor}: The patient's history of gastric ulcers, combined with the sudden onset of severe upper abdominal pain, nausea, and vomiting, suggests a possible acute gastrointestinal issue. Given the absence of new medications, stress, or lifestyle changes, we should consider complications related to the existing gastric ulcers or other acute conditions like pancreatitis or cholecystitis.
To further narrow down the diagnosis, we need to review the results of any recent medical examinations or tests.
4. Have you undergone any recent medical examinations or tests, such as blood tests, abdominal ultrasound, or CT scans?\\

\end{tabularx}}

\caption{The case study about consultation with one-shot CoT, the doctor model is GPT-4o. The $\rm <image.jpg>$ will be replaced with an image in the practice.}
\label{tab:case3-1}
\end{table*}

\begin{table*}[h]
\renewcommand{\arraystretch}{1.0}
\centering
\resizebox{16cm}{!}{
\begin{tabularx}{\textwidth}{X}
\rowcolor[HTML]{343434} 
{\color[HTML]{FFFDFA} The case study for one-shot CoT (B)}   
\\
\textbf{Patient}: 
\begin{enumerate}
    \item Blood tests show: WBC 2.46*10$^9$/L, Neu\% 79.0\%, Lym\% 14.0\%, Neu$\#$ 1.95*10$^9$/L, Lym$\#$ 0.34*10$^9$/L, Hb 159.00 g/L. Protein 45.7 g/L, Albumin 22.9 g/L, Globulin 23.60 g/L.
    \item Abdominal ultrasound: Everything is normal.
    \item CT scans: $\rm <image.jpg>$.
\end{enumerate}

\textbf{Doctor}: Based on the provided CT scan image, it appears there is a significant finding in the upper abdomen. The image shows signs consistent with acute pancreatitis, which can present with severe upper abdominal pain, nausea, and vomiting. The blood test results also show an elevated neutrophil percentage, which can indicate an inflammatory process.

Given the patient's history of gastric ulcers and the current presentation, acute pancreatitis is a likely diagnosis. This condition requires prompt medical attention and management, including fasting, IV fluids, pain management, and monitoring for complications.

\textbf{Preliminary Diagnosis: Acute Pancreatitis}

\textbf{Recommendations}:
\begin{enumerate}
    \item Hospital admission for close monitoring and supportive care.
    \item Fasting to rest the pancreas.
    \item Intravenous fluids to maintain hydration.
    \item Pain management with appropriate analgesics.
    \item Monitoring for potential complications such as infection, necrosis, or organ failure.
    \item Further evaluation by a gastroenterologist for comprehensive management.
\end{enumerate}

Please proceed to the hospital immediately for further evaluation and treatment. \\ \hline
\textbf{Medical Report} \\

\textbf{Patient Information:}
\begin{itemize}
    \item \textbf{Name:} [Patient's Name]
    \item \textbf{Age:} 66 years
    \item \textbf{Gender:} Male
    \item \textbf{Occupation:} Farmer
\end{itemize}

\textbf{Present Illness:}
The patient presented with a sudden onset of severe upper abdominal pain lasting one day. The pain is constant and localized to the upper abdomen, without radiation to other parts of the body. The patient also reported experiencing nausea and vomiting six times, with the vomitus consisting of gastric contents totaling about 1000 milliliters, containing no blood or worms. There is no mention of fever or changes in bowel movements.

\textbf{Past Medical History:}
\begin{itemize}
    \item History of gastric ulcers for three years, intermittently treated with oral medications.
    \item No history of gallstones or pancreatitis.
    \item No recent intake of new medications, significant stress, or changes in diet or lifestyle.
\end{itemize}

\textbf{Examination Items and Results:}
\begin{enumerate}
    \item \textbf{Blood Tests:} WBC: 2.46$\times$10$^9$/L, Neu$\%$: 79.0$\%$, Lym$\%$: 14.0$\%$,  Neu$\#$: 1.95$\times$10$^9$/L, Lym$\#$: 0.34$\times$10$^9$/L,  Hb: 159.00 g/L, Protein: 45.7 g/L, Albumin: 22.9 g/L, Globulin: 23.60 g/L
    
    \item \textbf{Abdominal Ultrasound:} Normal findings.

    \item \textbf{CT Scan:}  The CT scan image shows signs consistent with acute pancreatitis.
\end{enumerate}

\textbf{Diagnosis:}
\begin{itemize}
    \item \textbf{Preliminary Diagnosis:} Acute Pancreatitis
\end{itemize}

\textbf{Treatment Plan:}
\begin{enumerate}
    \item \textbf{Hospital Admission:}  Immediate hospital admission for close monitoring and supportive care. 
    
    \item \textbf{Fasting:} Fasting to rest the pancreas and reduce pancreatic enzyme secretion. 
    
    \item \textbf{Intravenous Fluids:} Administration of IV fluids to maintain hydration and electrolyte balance. 
    
    \item \textbf{Pain Management:} se of appropriate analgesics to manage pain. 
    
    \item \textbf{Monitoring for Complications:} Close monitoring for potential complications such as infection, necrosis, or organ failure. 
    
    \item \textbf{Further Evaluation:}  Referral to a gastroenterologist for comprehensive management and further evaluation. 
\end{enumerate}

\textbf{Recommendations:}
The patient is advised to proceed to the hospital immediately for further evaluation and treatment. Prompt medical attention is necessary to manage acute pancreatitis and prevent potential complications.

\textbf{Follow-Up:}
\begin{itemize}
    \item Regular follow-up with a gastroenterologist to monitor the patient's condition and adjust the treatment plan as needed.
    \item Additional imaging studies and laboratory tests may be required to assess the progression of the disease and response to treatment.
\end{itemize}

\textbf{Conclusion:}
The patient presents with symptoms and imaging findings consistent with acute pancreatitis. Immediate medical intervention is essential to manage the condition and prevent complications. The patient should be admitted to the hospital for comprehensive care and monitoring.

\end{tabularx}}

\caption{The case study about consultation with one-shot CoT, the doctor model is GPT-4o. The $\rm <image.jpg>$ will be replaced with an image in the practice.}
\label{tab:case3-2}
\end{table*}


\end{document}